\documentclass[10pt,twocolumn,letterpaper]{article}

\usepackage{iccv}
\usepackage{times}
\usepackage{epsfig}
\usepackage{graphicx}
\usepackage{amsmath}
\usepackage{amssymb}

\usepackage[ruled]{algorithm2e}
\usepackage{bbm}
\usepackage{bm}
\usepackage{multirow}
\usepackage{array}
\usepackage{booktabs}
\usepackage{diagbox}
\usepackage[table]{xcolor}
\usepackage{bbding}

\newcommand{\Tref}[1]{Table~\ref{#1}}

\newcommand{\Fref}[1]{Figure~\ref{#1}}
\newcommand{\Sref}[1]{Section~\ref{#1}}

\newcommand{\sref}[1]{Sec.~\ref{#1}}

\usepackage[pagebackref=true,breaklinks=true,letterpaper=true,colorlinks,bookmarks=false]{hyperref}
\usepackage[accsupp]{axessibility}  

\iccvfinalcopy 

\def\iccvPaperID{2368} 

\ificcvfinal\fi

\begin{document}

\title{Improving Adversarial Robustness of Masked Autoencoders via\\ Test-time Frequency-domain Prompting}


\author{
Qidong Huang\textsuperscript{\rm 1} \quad
Xiaoyi Dong\textsuperscript{\rm 2} \quad
Dongdong Chen\textsuperscript{\rm 3} \quad
Yinpeng Chen\textsuperscript{\rm 3} \quad
Lu Yuan\textsuperscript{\rm 3} \quad \\
Gang Hua\textsuperscript{\rm 4} \quad
Weiming Zhang\textsuperscript{\rm 1,}\thanks{Corresponding author.} \quad
Nenghai Yu\textsuperscript{\rm 1} \\
\textsuperscript{\rm 1}University of Science and Technology of China\quad \
\textsuperscript{\rm 2}Shanghai AI Lab \\ 
\textsuperscript{\rm 3}Microsoft Reaserch\quad \ 
\textsuperscript{\rm 4}Wormpex AI Research \ \\
{\tt\small\{hqd0037@mail., zhangwm@, ynh@\}ustc.edu.cn}\quad
{\tt\small dongxiaoyi@pjlab.org.cn}
\\
{\tt\small\{cddlyf@, ganghua@\}gmail.com}\quad
{\tt\small\{yiche@, luyuan@\}microsoft.com}
}

\maketitle
\ificcvfinal\thispagestyle{empty}\fi

\begin{abstract}
In this paper, we investigate the adversarial robustness of vision transformers that are equipped with BERT pretraining (\eg, BEiT, MAE). 
A surprising observation is that MAE has significantly worse adversarial robustness than other BERT pretraining methods. 
This observation drives us to rethink the basic differences between these BERT pretraining methods and how these differences affect the robustness against adversarial perturbations. 
Our empirical analysis reveals that the adversarial robustness of BERT pretraining is highly related to the reconstruction target, \ie, predicting the raw pixels of masked image patches will degrade more adversarial robustness of the model than predicting the semantic context, since it guides the model to concentrate more on medium-/high-frequency components of images. 
Based on our analysis, we provide a simple yet effective way to boost the adversarial robustness of MAE. 
The basic idea is using the dataset-extracted domain knowledge to occupy the medium-/high-frequency of images, thus narrowing the optimization space of adversarial perturbations. 
Specifically, we group the distribution of pretraining data and optimize a set of cluster-specific visual prompts on frequency domain. 
These prompts are incorporated with input images through prototype-based prompt selection during test period. 
Extensive evaluation shows that our method clearly boost MAE's adversarial robustness while maintaining its clean performance on ImageNet-1k classification. 
Our code is available at: \href{https://github.com/shikiw/RobustMAE}{https://github.com/shikiw/RobustMAE}. 
\end{abstract}

\section{Introduction}

BERT pretraining for vision transformers \cite{beit21,mae21,peco21,ibot21,dong2023maskclip} ignites rising interests recently and shows great potentials in improving downstream performance. Existing methods share the similar pretraining principle, \ie, randomly masking some patches on the input images and asking the transformer to reconstruct them based on unmasked patches, which encourages the transformer model to comprehend the semantic relationship among different patches for better reconstruction. Thus their main differences lie in the reconstruction targets. 

\begin{figure}[t]
\centering
\includegraphics[width=1.0\linewidth]{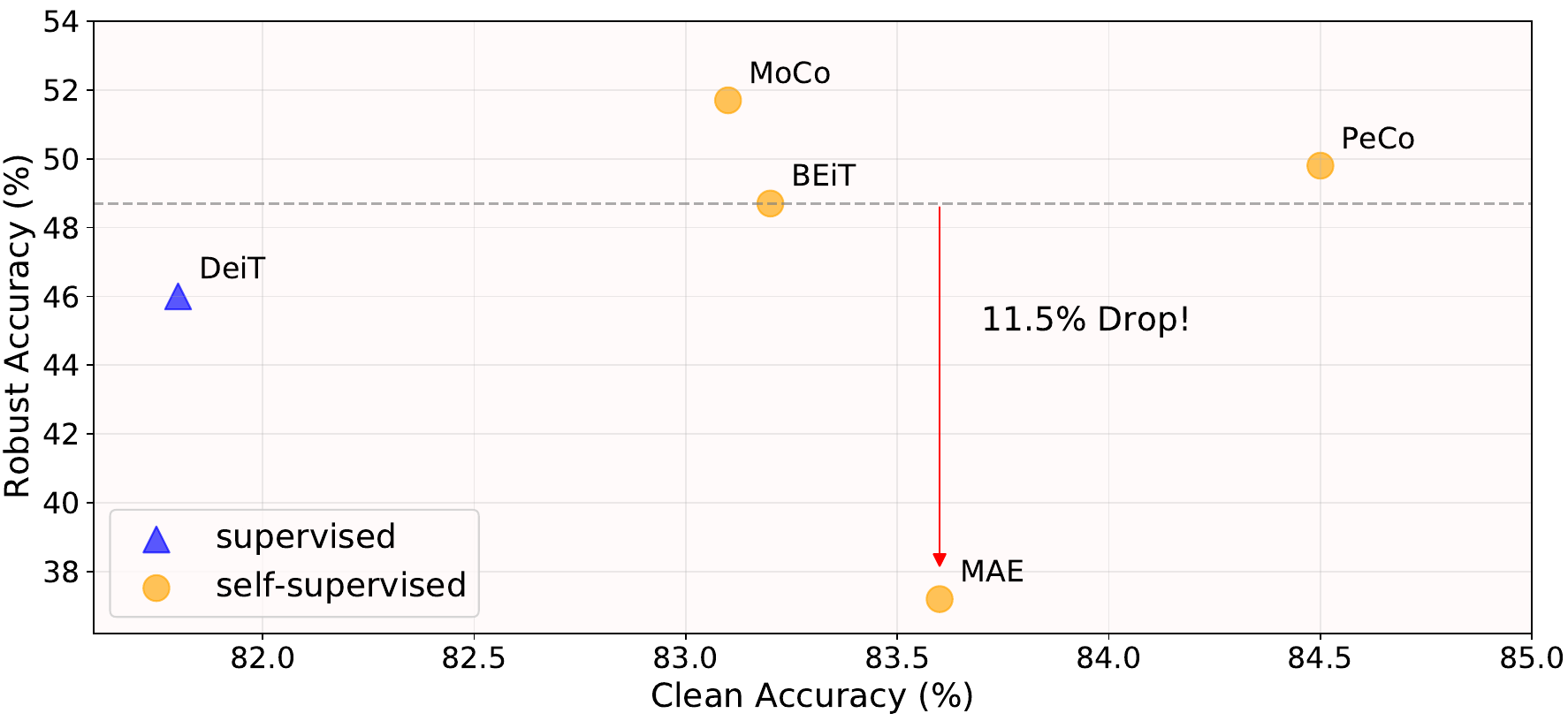}
\caption{\textbf{MAE has much worse adversarial robustness than other pretraining counterparts.} We use ViT-Base as the backbone and evaluate on ImageNet-1k. The robust accuracy is tested under standard $l_\infty$-norm PGD attack.}
\label{fig:teaser}
\end{figure}

Specifically, following the practices in natural language processing \cite{DevlinCLT19}, BEiT \cite{beit21} and PeCo \cite{peco21} convert the image patches into discrete tokens with pretrained VQVAE \cite{van2017neural} and regard such discrete tokens as the prediction targets. iBOT \cite{ibot21} integrates contrastive learning and uses online learned feature as prediction target. MAE \cite{mae21} takes a step forward to directly use the raw pixel values as the target, achieving surprisingly good results. It can be regarded as a special form of denoising autoencoding, which provides a possibility to revitalize this classical direction. Although these methods have different reconstruction targets, they all achieve substantial improvement over multiple vision tasks. 


However, we observe a surprising discrepancy between MAE and other pretraining counterparts (\eg, BEiT, PeCo) on \textit{adversarial robustness}. As shown in \Fref{fig:teaser}, MAE is significantly more vulnerable to standard $l_\infty$-norm PGD \cite{Madry18adversarial} adversarial attack than BEiT and PeCo (drop 11.5\% in ImageNet-1k top-1 accuracy). This paper is motivated by this observation to explore the causing factors, which naturally lead to our solution.


Our analysis manifests that MAE has the stronger dependence on medium-/high-frequency signals of images, thus it is more sensitive to the medium-/high-frequency noise introduced by regular adversarial attacks. This is exactly due to the difference between MAE and its counterparts (BEiT, PeCo) in reconstruction task of pretraining. Specifically, MAE reconstructs raw pixels of input images, while BEiT and PeCo reconstructs the semantic class for each image patch. Predicting raw pixels drives MAE to rely more on medium-/high-frequency information of input images. To validate this, we filter out both medium-/high-frequency components of clean images with a low pass filter and perform the evaluation. MAE suffers significantly more accuracy degradation than other pretraining methods (Please refer to \Fref{fig:fsm} and \ref{fig:freq_acc}).

Based on the above observations, we propose a simple yet effective solution to improve the adversarial robustness of MAE. Our goal is to learn a set of frequency-domain visual prompts to enhance the medium-/high-frequency signals of input images during test time. In detail, we first analyze the representations learned by the pretrained model and then partition it into several clusters. On the samples distributed in each cluster, a visual prompt are optimized in frequency domain by minimizing the classification objective loss, of which the generation is constrained on medium-/high-frequency components of images.
It naturally fills the medium-/high-frequency components with the dataset-extracted patterns, so that searching regular adversarial perturbations on these components can be more difficult. 
Experiments tells that the proposed solution not only significantly boost the adversarial robustness of MAE to surpass other pretraining counterparts, but also generally maintain the clean performance on the ImageNet-1k dataset.

\section{Preliminaries}

\subsection{Vision BERT Pretraining}
\label{sec:2.1}
BERT pretraining \cite{DevlinCLT19} has become the dominant self-supervised learning mechanism in the field of natural language processing (NLP). The key idea is masked language modeling (MLM), \ie, randomly masking some words in the sentence and asking the transformer to reconstruct them based on the remaining unmasked words. This reconstruction process requires the transformer model to understand underlying semantics and model the relationship between unmasked words and masked words, thus helping learn meaningful representations.

\begin{figure}[t]
\centering
\includegraphics[width=\linewidth]{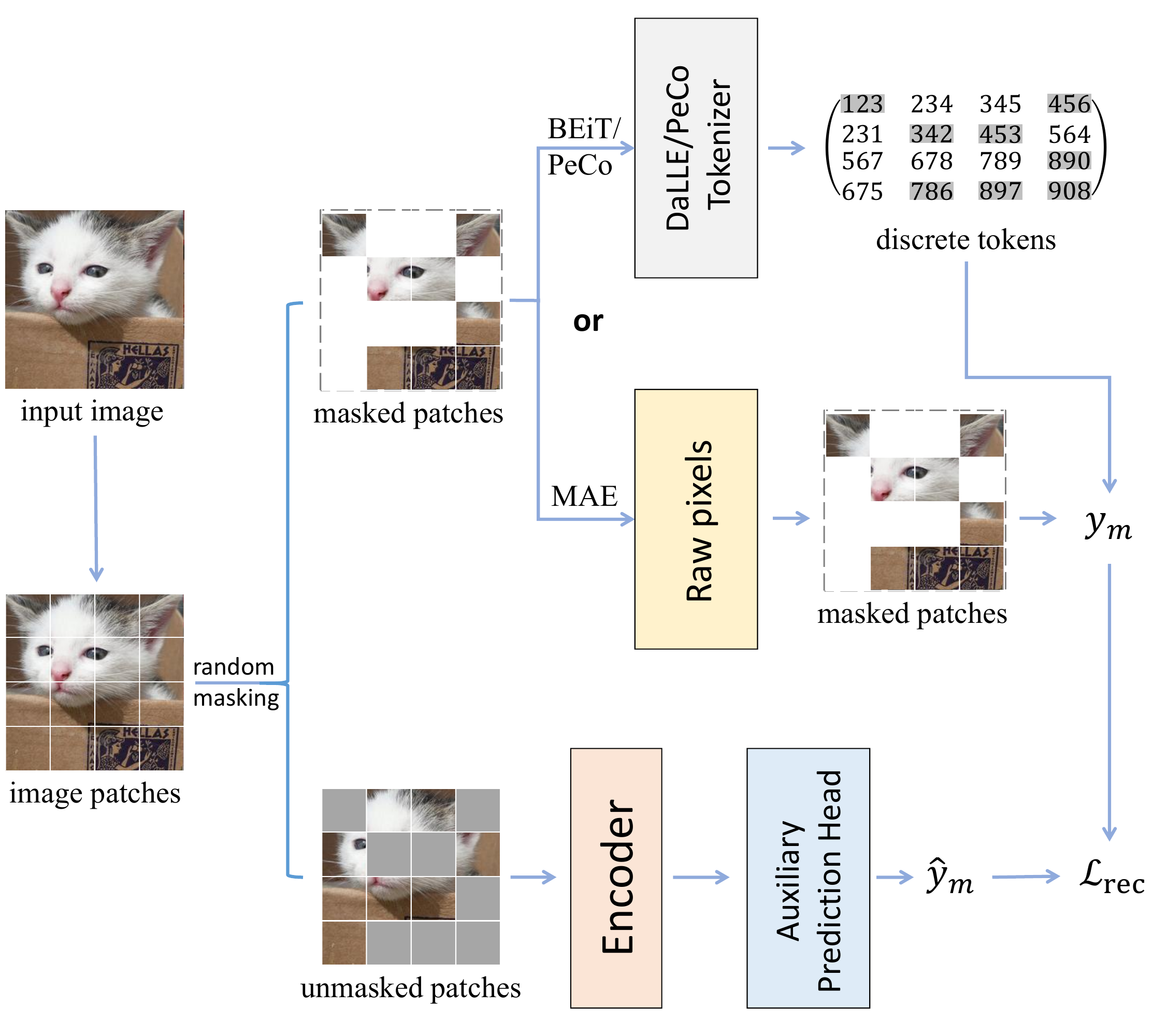}
\caption{Key difference among BEiT, PeCo and MAE, where MAE directly use the original image patches for pixel-level reconstruction, BEiT and PeCo choose the discrete tokens as the target to reconstruct semantic context.}
\label{fig:bert}
\end{figure}

The recent surge of vision BERT pretraining follows the similar logic. As illustrated in \Fref{fig:bert}, each input image is converted into a series of patches (tokens) and some patches are randomly masked out. The unmasked patches $\mathbf{x}_u$ are fed into the transformer encoder $\mathbf{E}$, on top of which an auxiliary prediction head $\mathbf{h}$ is used to reconstruct masked patches $\mathbf{x}_m$ by minimizing reconstruction loss $\mathcal{L}_{rec}$. Formally, 

\begin{equation}
\begin{aligned}
    \hat{\mathbf{y}}_m &= \mathbf{h}(\mathbf{E}(\mathbf{x}_u, \mathbf{p}_{m,u}), \mathbf{p}_{m,u}), \\
    \mathcal{L}_{rec} &= L(\hat{\mathbf{y}}_m, \mathbf{y}_m),
\end{aligned}
\end{equation}
where $\mathbf{p}_{m,u}$ indicates the positional information of masked and unmasked tokens, thus the transformer model comnprehends the spatial relationship of tokens. $\hat{\mathbf{y}}_m, \mathbf{y}_m$ are the predicted and ground-truth reconstruction representations of masked tokens, respectively. Depending on the format of $\mathbf{y}$, $L$ is the corresponding reconstruction loss function (\eg, cross-entropy loss in BEiT/PeCo and $\ell_2$ loss in MAE). By analogy, the above vision BERT pretraining strategy is often called ``masked image modeling (MIM)''. 

The main difference among existing vision BERT pretraining methods lies in the design of $\mathbf{y}$. More specifically, BEiT \cite{beit21} directly takes the pretrained DaLLE's VQVAE tokenizer to convert each image patch into discrete tokens (indexed in a large codebook), where $\mathbf{y}$ is the code index of each image patch. PeCo \cite{peco21} follows the similar token design and designs a new perceptual VQVAE tokenizer, striving to make the discrete tokens more semantic. MaskFeat~\cite{maskfeat21} extracts the lowe-level HOG features of image patches as $\mathbf{y}$. MAE\cite{mae21} takes a big step forward, which directly uses the normalized raw pixel values as $\mathbf{y}$, \ie, $\mathbf{y}_m=norm(\mathbf{x}_m)$. In the following sections, we will show that the design difference of $\mathbf{y}$ is a key impact factor that influences the adversarial robustness of different BERT pretraining models.

After BERT pretraining, the pretrained transformers will be finetuned on downstream vision tasks, including but not limited to image classification, object detection and semantic segmentation. Considering it is difficult to evaluate the adversarial robustness of the pretrained model itself, we follow the common practice \cite{JiangCCW20,paul2022vision,mao2022towards,shao2021adversarial,naseerimproving} and instead test the the supervised finetuned model to evaluate the pretraining robustness. 

\begin{figure*}[t]
\vspace{-1em}
\begin{minipage}{0.32\linewidth}
    \centering
    \includegraphics[width=1\linewidth]{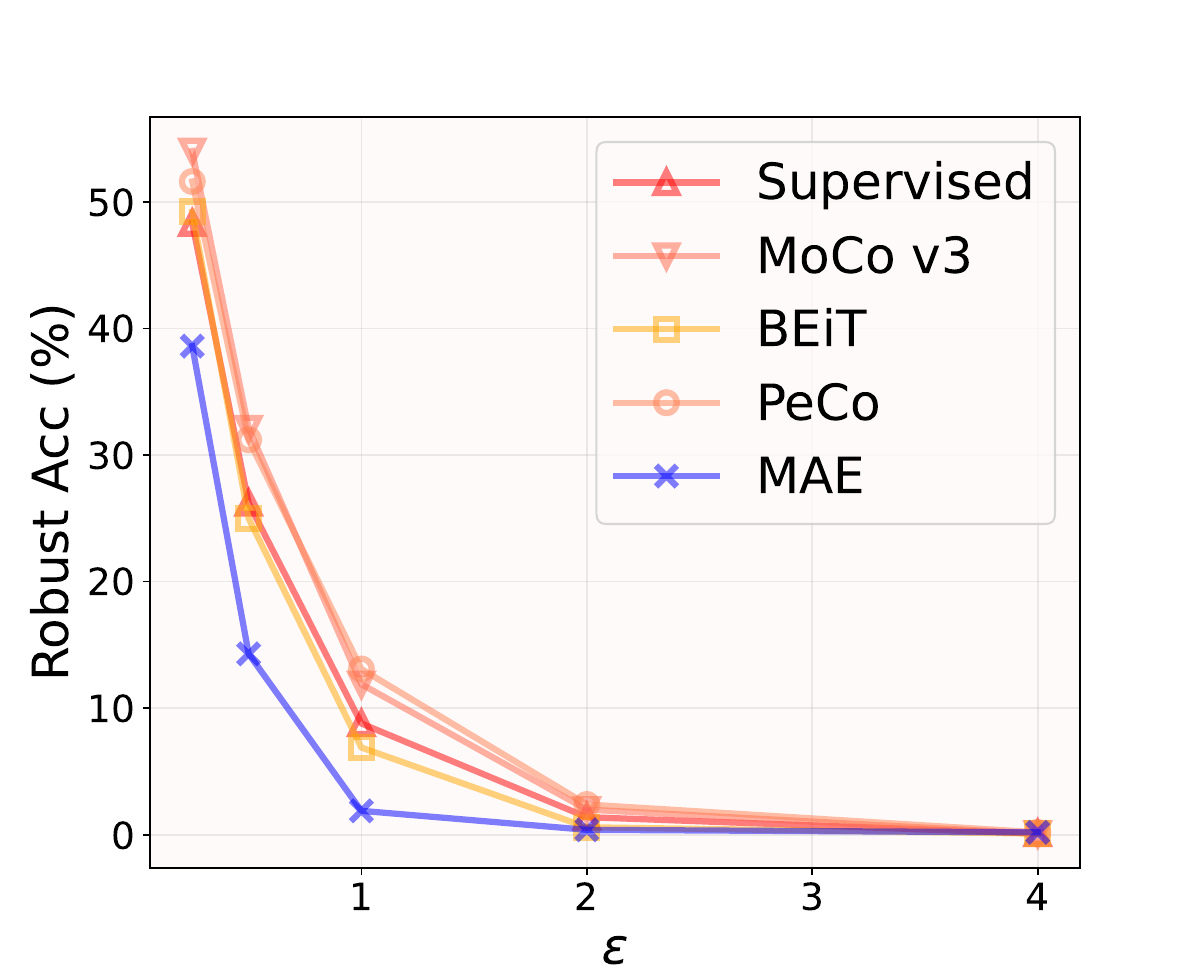}
    \\ 
    \footnotesize PGD
\end{minipage}
\hfill
\begin{minipage}{0.32\linewidth}
    \centering
    \includegraphics[width=1\linewidth]{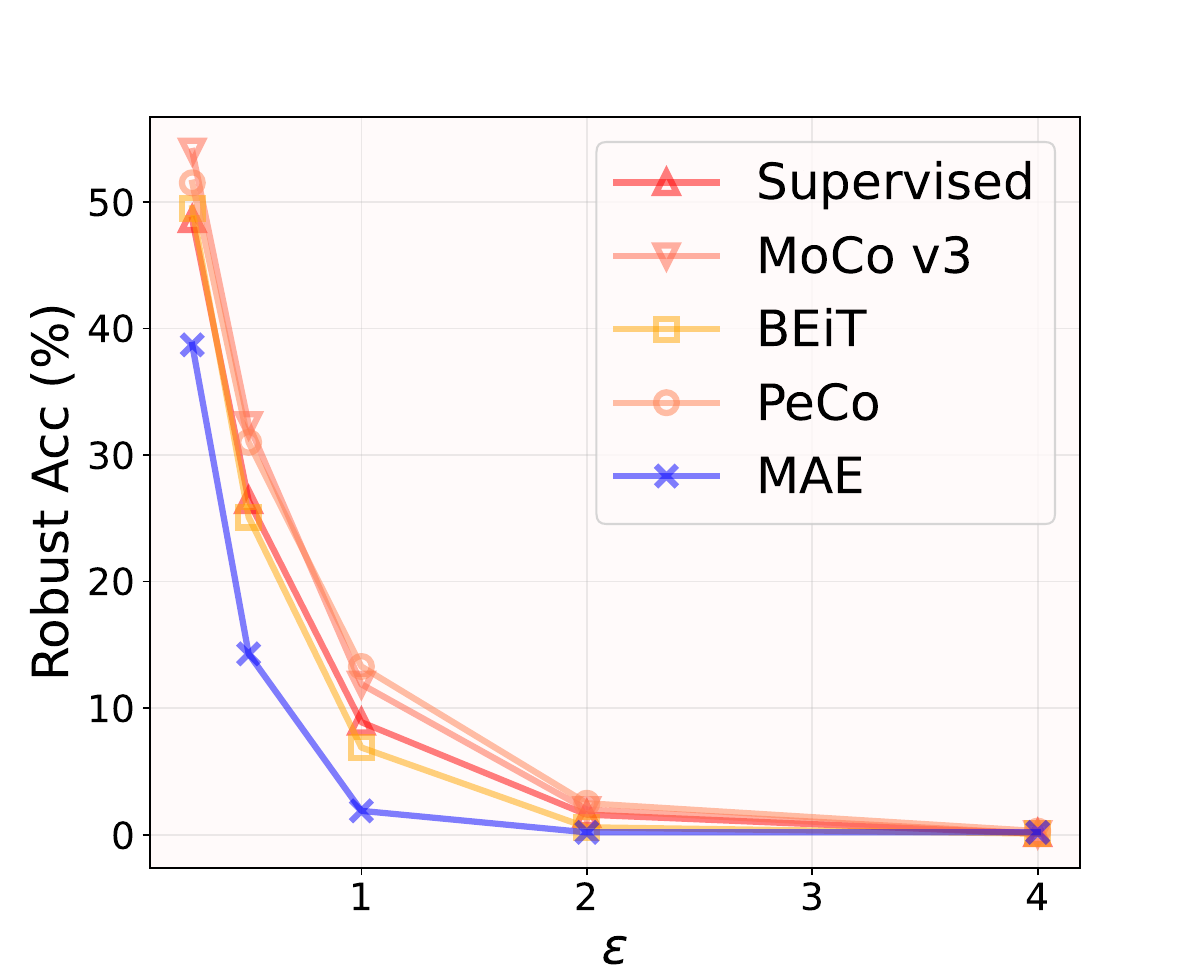}
    \\ 
    \footnotesize BIM
\end{minipage}
\hfill
\begin{minipage}{0.32\linewidth}
    \centering
    \includegraphics[width=1\linewidth]{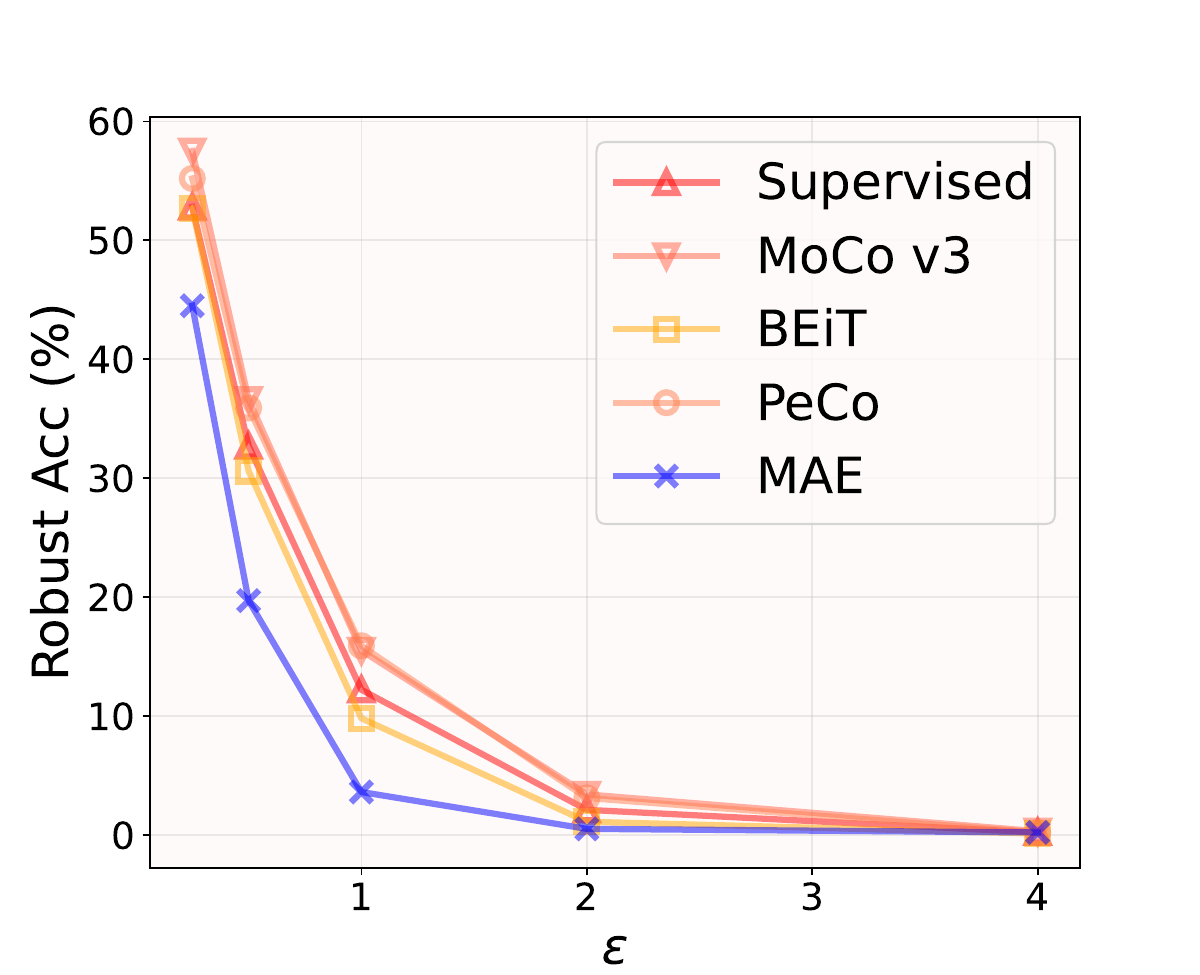}
    \\
    \footnotesize MIM
\end{minipage}
\vspace{0.5em}
\caption{Comparison of adversarial robustness to various adversarial attacks ($\epsilon = \varepsilon/255$) among different BERT pretraining methods. All methods are equally implemented on ViT-B/16 backbone. \textbf{MAE shows significantly worse adversarial robustness than other methods.}}
\label{fig:adv_curve}
\end{figure*}

\begin{table*}[t]
	\footnotesize
    \centering
    \setlength{\tabcolsep}{0.35mm}{
    \begin{tabular}{lc|p{12mm}<{\centering}p{12mm}<{\centering}p{12mm}<{\centering}p{12mm}<{\centering}|p{12mm}<{\centering}p{12mm}<{\centering}p{12mm}<{\centering}p{12mm}<{\centering}|p{12mm}<{\centering}}
        \toprule
        \multicolumn{1}{c}{\multirow{2}{*}{Method}}
        & \multicolumn{1}{c}{\multirow{2}{*}{Clean Acc (\%)}} 
        & \multicolumn{4}{|c}{Robust Acc (\%) ($\epsilon=0.5/255$)} 
        & \multicolumn{4}{|c}{Robust Acc (\%) ($\epsilon=1.0/255$)} 
        & \multicolumn{1}{|c}{ } 
        \\
        \cmidrule{3-6}
        \cmidrule{7-10}
        & 
        & PGD\cite{Madry18adversarial}
        & MIM\cite{dong18mifgm}
        & BIM\cite{kurakin2016adversarial}
        & AA\cite{croce2020autoattack}
        & PGD\cite{Madry18adversarial}
        & MIM\cite{dong18mifgm}
        & BIM\cite{kurakin2016adversarial}
        & AA\cite{croce2020autoattack}
        & C\&W\cite{carlini2017towards}
        \\
        \midrule
        Supervised & 81.8 & 26.2 & 32.7 & 26.4 & 6.1 & 8.8 & 12.2 & 8.9 & 0.1 & 18.6
        \\
        MoCo v3 \cite{Chen21mocov3} & 83.1 & 32.1 & 36.6 & 32.4 & 7.8 & 11.9 & 15.5 & 11.9 & 0.4 & 25.2
        \\
        BEiT \cite{beit21} & 83.2 & 25.0 & 30.6 & 25.1 & 4.5 & 6.9 & 9.8 & 6.9 & 0.0 & 21.6
        \\
        PeCo \cite{peco21} & 84.5 & 31.2 & 35.9 & 31.0 & 3.9 & 13.1 & 15.9 & 13.3 & 0.1 & 24.0
        \\
        MAE \cite{mae21} & 83.6 & \textbf{14.3} & \textbf{19.7} & \textbf{14.3} & \textbf{1.6} & \textbf{1.9} & \textbf{3.6} & \textbf{1.9} & \textbf{0.0} & \textbf{5.9}
        \\
        \bottomrule
    \end{tabular}
    }
    \vspace{0.5em}
    \caption{Comparison of adversarial robustness to various adversarial attacks among different BERT training methods.  All methods are equally implemented on ViT-B/16 backbone, where MAE shows significantly worse robustness than other methods. Here ``Clean Acc" and ``Robust Acc" denote the top-1 classification accuracy on clean samples and adversarial samples, respectively.}
    \label{tab:table1}
\end{table*}

\subsection{Adversarial Attack and Defense} 

Deep neural networks (DNNs) are vulnerable to adversarial samples \cite{SzegedyZSBEGF13,goodfellow2015explaining,wei2022towards,qin2021understanding,tang2021robustart,huang2022siadv,dong2020robust,zhou2020lg,dong2020greedyfool}, which are generated by adding imperceptible perturbations on clean inputs to mislead the model prediction. 
Given a classification model $\mathcal{C} (x)$, the goal of adversarial attack is to search for a slight perturbation under the strict constraint until $\mathcal{C}$ has been fooled. 
Usually, for an input $x$ with its ground-truth label $y$, the perturbation added on the generated adversarial sample $x'$ is constrained within $l_p$-norm ball, \ie, $\parallel x-x' \parallel_p \leq \epsilon$.  Typically, adversarial attacks can be divided into two categories: 
1) Targeted attack that makes $\mathcal{C} (x') = y'$ where $y' \neq y$ is the label specified by the attacker. 
2) Untargeted attack that just require $x'$ to mislead the model $\mathcal{C}$ to give wrong decision $\mathcal{C} (x') \neq y$ without specifying the attack label. The adversarial robustness of one model is defined as whether this model can still correctly classify adversarial samples under the predefined attack strength, \ie, classification accuracy on input adversarial samples.

Many defense methods have been proposed to improve model robustness against adversarial attacks, such as introducing stochasticity \cite{DhillonALBKKA18,bian2021adversarial}, feature squeezing \cite{PapernotM17} or robust module designs \cite{dong2020self}. 
Serving as the most effective way to defend against adversarial attack, adversarial training does not need any change on model architecture and directly engages adversarial samples during model training. Representative adversarial training methods include static ensemble models \cite{TramerKPGBM18}, adversarial logit pairing and universal first-order adversary \cite{Madry18adversarial,zhang2019trades,cui2020lbgat} (\ie, using PGD as the inner loop).
Under the min-max optimization principle, adversarial training alternates between adversarial sample generation and model update, which follows the generic form
\begin{equation}
    \mathop{\text{min}}_\theta \mathbb{E}_{x\in \mathcal{D}} \big[ \mathop{\text{max}}_{x'} \ell (x';\theta) \big],\quad \textit{s.t.} \parallel x-x'\parallel_p \leq \epsilon
\end{equation}
where $\theta$ denotes the learnable parameters of the model, $\mathcal{D}$ denotes the training dataset, $\ell$ is the training objective loss and $\epsilon$ is the attack budget.

\section{Adversarial Robustness of MAE}
In this section, we prove that MAE has significant more degradation than other BERT pretraining methods when dealing with adversarial examples. \textit{Our analysis shows that MAE's sensitivity to adversarial noises is attributed to its strong dependence on medium-/high-frequency signals.} 


\begin{figure*}[t]
\begin{minipage}{0.19\linewidth}
    \centering
    \includegraphics[width=1\linewidth]{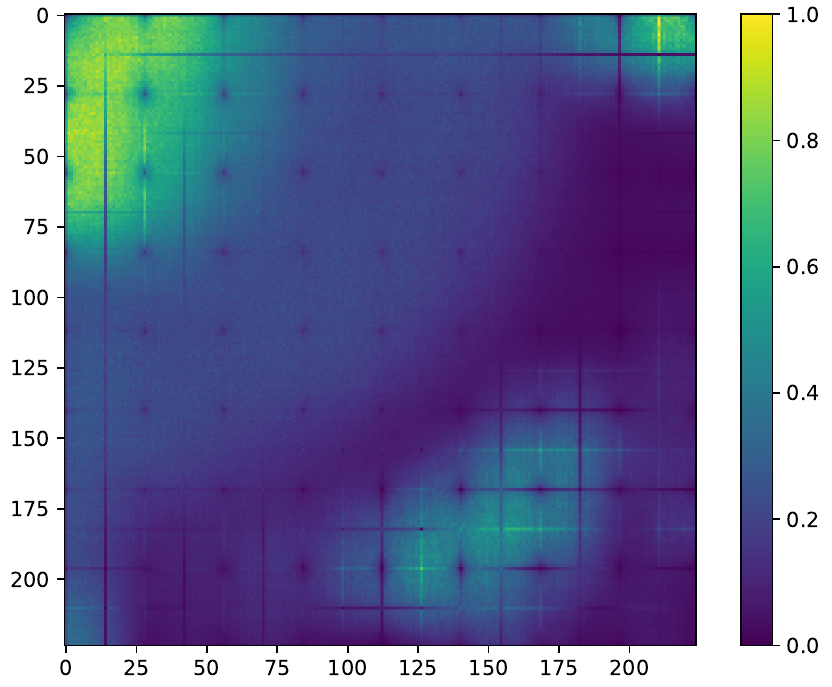}
    \\ 
    \footnotesize Supervised
\end{minipage}
\hfill
\begin{minipage}{0.19\linewidth}
    \centering
    \includegraphics[width=1\linewidth]{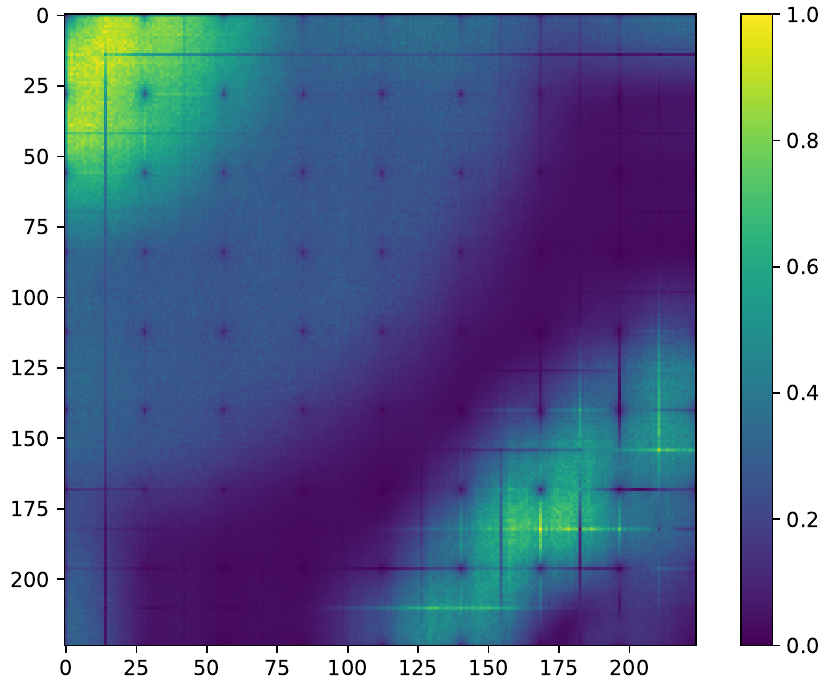}
    \\ 
    \footnotesize MoCo v3
\end{minipage}
\hfill
\begin{minipage}{0.19\linewidth}
    \centering
    \includegraphics[width=1\linewidth]{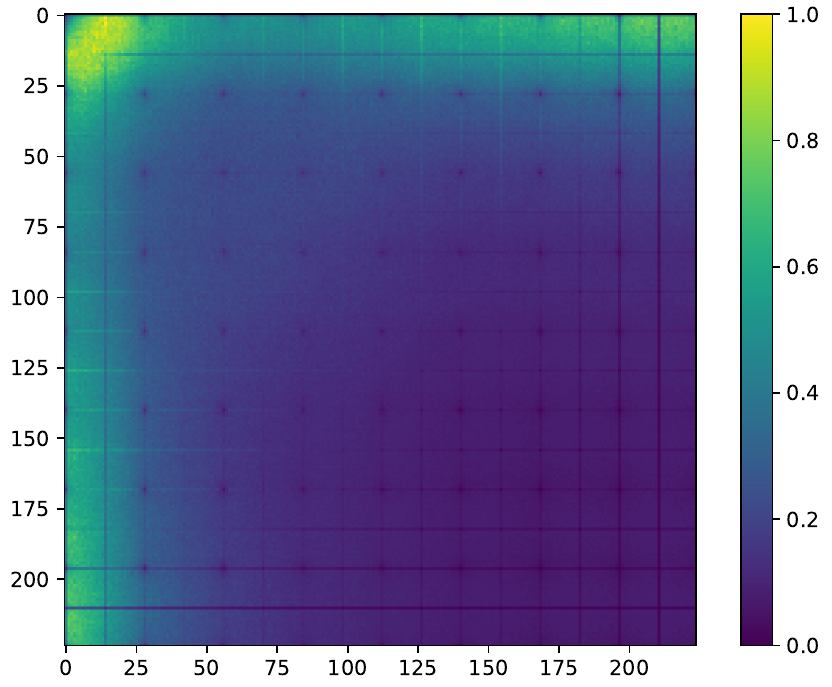}
    \\
    \footnotesize BEiT
\end{minipage}
\hfill
\begin{minipage}{0.19\linewidth}
    \centering
    \includegraphics[width=1\linewidth]{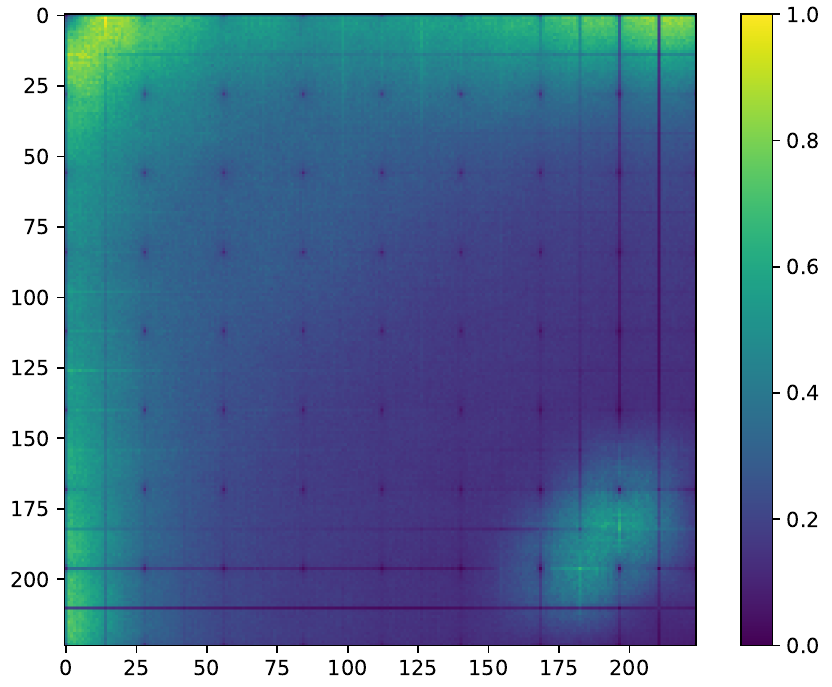}
    \\
    \footnotesize PeCo
\end{minipage}
\hfill
\begin{minipage}{0.19\linewidth}
    \centering
    \includegraphics[width=1\linewidth]{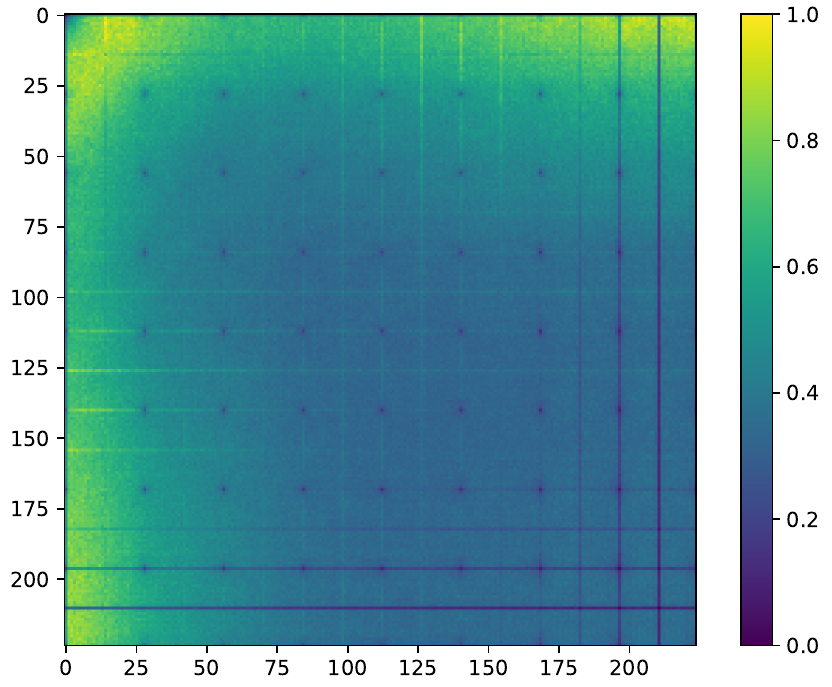}
    \\
    \footnotesize MAE
\end{minipage}
\vspace{0.5em}
\caption{Frequency saliency maps of different pretrained models based on Discrete Cosine Transform (DCT). From the top left corner to the bottom right corner are low, medium and high frequencies. The brighter colors denote the greater sensitivity on the corresponding frequency component. \textbf{MAE shows the relatively stronger dependence on medium-/high-frequency components.}}
\label{fig:fsm}
\end{figure*}

\subsection{MAE's Degradation on Adversarial Robustness}
\label{sec:3.1}

We first verify the robust accuracy of different pretrained models on ImageNet-1k \cite{DengDSLL009} classification. 
Five types of adversarial attacks, including PGD \cite{Madry18adversarial}, BIM \cite{kurakin2016adversarial}, MIM \cite{dong18mifgm}, C\&W \cite{carlini2017towards} and Auto-Attack (AA) \cite{croce2020autoattack}, are selected here for the robustness evaluation, where PGD, BIM and MIM are implemented for 20 iterations, with different attack budget $\epsilon$ under $l_\infty$-norm constraint (The $l_2$-norm results can be found in our supplementary). 
For fair comparison, all methods equally employ the same ViT-B/16 \cite{DosovitskiyB0WZ21} backbone, and all of BERT pretraining methods are conducted for 800 epochs, sharing the same training recipe. 
The pretrained models are finetuned in a supervised manner on ImageNet-1k for 100 epochs. 
The adversarial robustness we study here concentrates on the finetuned models.

From the quantitative results listed in \Tref{tab:table1} and \Fref{fig:adv_curve}, we can surprisingly find that, though MAE \cite{mae21} pretrained model performs better or comparably on clean images, it encounters much larger performance drop on adversarial samples. For example, in \Tref{tab:table1}, the robust accuracy of MAE is significantly lower than the supervised baseline, \ie, 11.9\% lower on PGD, 13.0\% lower on MIM, 12.7\% lower on C\&W and 4.5\% lower on Auto-Attack. But as for BEiT \cite{beit21} and PeCo \cite{peco21}, both of them can obtain comparable performance with the supervised pretraining baseline by contrast. For the self-supervised contrastive pretraining MoCo v3 \cite{Chen21mocov3}, it has slightly better robustness than all BERT pretraining methods but lower performance on clean samples. The better robustness should be relevant with the strong augmentation strategies used in contrastive-style pretraining, which are not required or considered in vision BERT pretraining. To summarize, though also adopting the BERT-style pre-text task during pretraining, MAE has much weaker ability to defend against adversarial attack compared with BEiT and PeCo. 
We give the further analysis in \sref{sec:3.2} to discuss the possible reason that causes MAE's degradation on adversarial robustness.

\subsection{Causing Factor Analysis}
\label{sec:3.2}

To clarify the reason, we start with the difference among these three methods in their implementation.
As illustrated in \sref{sec:2.1}, it is clear that the reconstruction target of MAE is significantly different from BEiT or PeCo. 
Specifically, MAE chooses the raw pixel values as the BERT pretraining prediction target for masked patches, where the simple mean square error loss on pixels is devised as the objective loss. 
Differently, the prediction target of BEiT or PeCo are discrete visual tokens that are encoded and quantized by a pretrained VQVAE, so these kind of targets are more semantic and information-denser than raw pixels. But how does the reconstruction target affects the adversarial robustness of pretrained models?

\vspace{0.5em}
\noindent\textbf{Assumption 1.} 
\textit{Predicting raw pixels of images makes MAE model more sensitive to medium-/high-frequency signals.}
\vspace{0.5em}

It has been recently proved that Convolutional Neural Networks (CNNs) shows the strong ability \cite{wang2020high} to perceive the high-frequency components of images that are almost imperceptible to human vision. This property, however, can be hardly observed in vision transformers (ViTs), since ViTs show less effective to capture the high-frequency components of images \cite{park2022vision,shao2021adversarial,bai2022improving,naseer2021intriguing}. In addition to the model architecture design, we hypothesis that the pretraining paradigm is also a crucial factor that affects the model sensitivity to the frequency-domain signals.

\begin{figure}[t]
\centering
\includegraphics[width=1.0\linewidth]{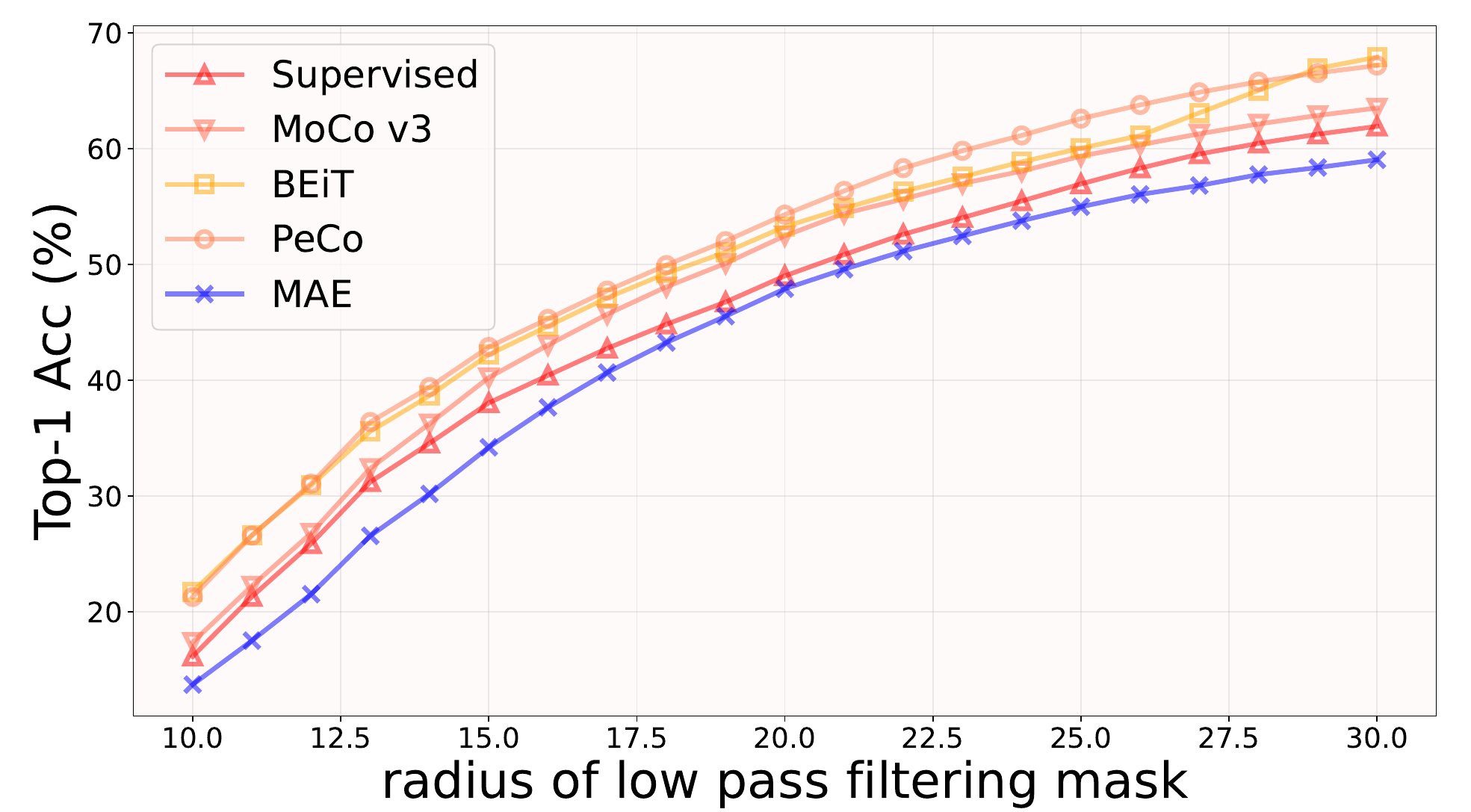}
\caption{Top-1 accuracy comparison on images that are filtered by the low pass frequency mask among different pretraining methods. \textbf{MAE is apparently more sensitive to the removed medium-/high-frequency signal in input images.}}
\label{fig:freq_acc}
\end{figure}

\begin{figure}[t]
\centering
\includegraphics[width=1.0\linewidth]{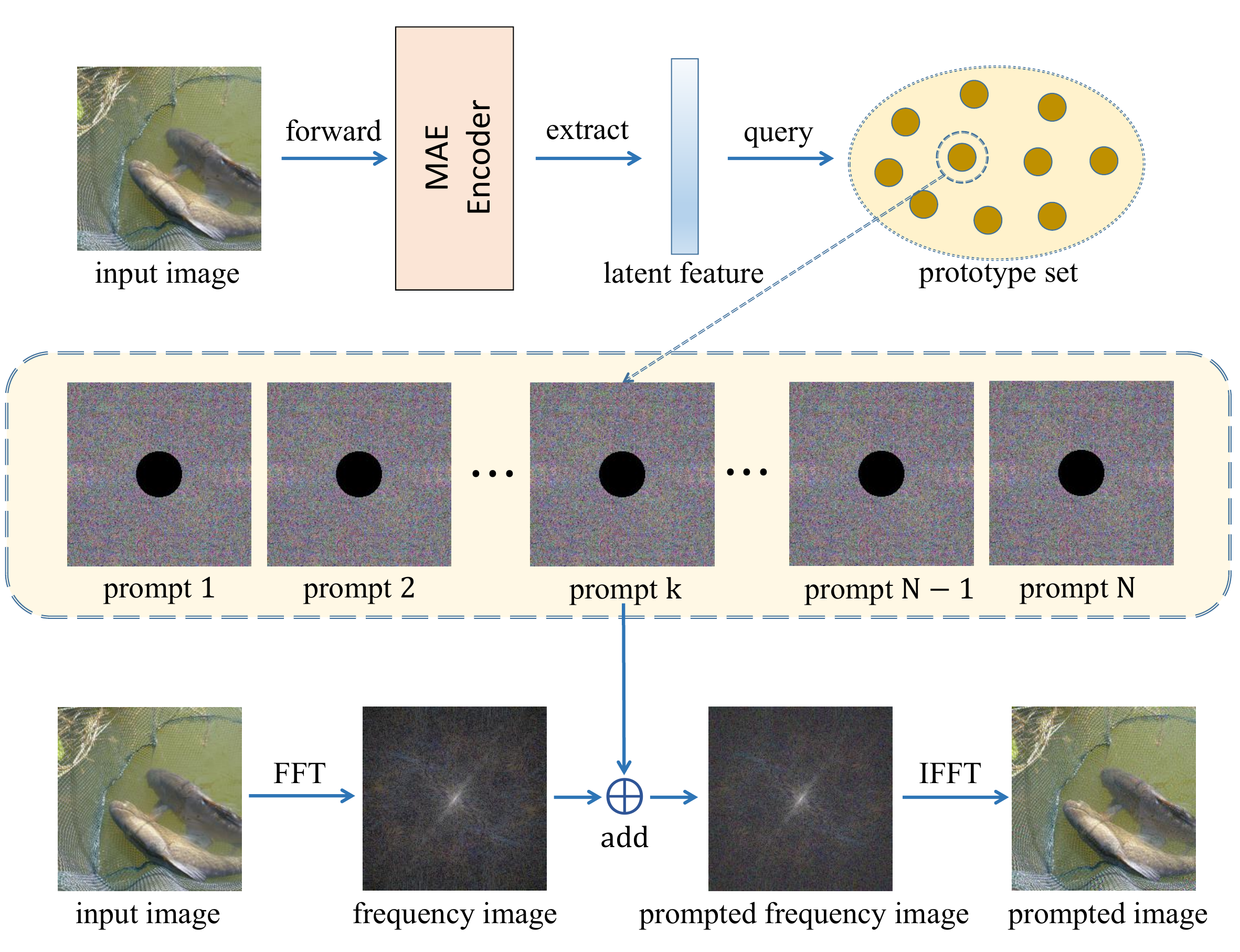}
\caption{The pipeline of our frequency-domain prompting to boost the adversarial robustness of MAE during test time.}
\label{fig:ours_pipeline}
\end{figure}

Following this clue, we conduct the experiment to study the classification performance of different BERT pretraining models on filtered images. These images are randomly sampled from the validation set of ImageNet-1k. We first try to remove the high-frequency components of images, where the images are converted into Fast Fourier Transform (FFT) space and only the central area (low frequency) is preserved with low pass filters of different sizes (similar to the mask used in \Fref{fig:ours_pipeline}). 
\Fref{fig:freq_acc} presents the relationship between top-1 classification accuracy and the radius used for low pass masking, from which we can observe that MAE's performance is worse than other BERT pretraining methods and the supervised baseline. In other words, when the image discards its medium-/high-frequency information, it is more difficult for MAE to capture any discriminative representation from the image to support its recognition. It reveals that MAE relies relatively less on low-frequency components of images and behaves more sensitive to the removing of medium-/high-frequency signals.

We further visualize the frequency sensitivity of these BERT pretraining models. 
To this end, we convert images into the frequency spectrum with Discrete Cosine Transform (DCT) and directly compute the gradients on the spectrum through back-propagation (minimizing cross-entropy loss as the objective). Intuitively, the larger absolute value of the gradient on frequency domain indicates the greater sensitivity. We construct the gradient maps and visualize them with heatmaps in \Fref{fig:fsm}, where it can be obviously observed that MAE requires more medium-/high-frequency signals (lower right triangle) for its prediction.

Therefore, we hypothesize that directly predicting pixels will make the MAE model more sensitive to medium-/high-frequency signals in the images. This claim is established on a simple fact that, raw pixels not only contain the basic information regarding the main content of images (usually presented as the low-frequency signals), but also includes a lot of trivial details (\eg, local textures or natural noises) that are not necessary for someone to recognize or describe the images. Accordingly, fitting on pixel values requires the neural learner to expend its capacity on precisely ``remember'' such details, \ie, concentrating more on medium-/high-frequency signals that are relatively less helpful for image understanding.


On the other hand, though it is dataset-dependent, adversarial perturbation generated by regular attacks on natural images is generally a special kind of medium-/high-frequency signals. Through disrupting the medium-/high-frequency components of the input image, adversarial perturbation brings much more negative effects to MAE than other pretraining methods. Therefore, it is easier to fool MAE with adversarial samples, which is consistent with MAE's degradation we observed in Table \ref{tab:table1}.

\vspace{0.5em}
\noindent\textbf{Assumption 2.} 
\textit{Predicting raw pixels of images helps MAE model become a more homogeneous vision learner.}
\vspace{0.5em}

\begin{figure}[t]
\centering
\includegraphics[width=1.0\linewidth]{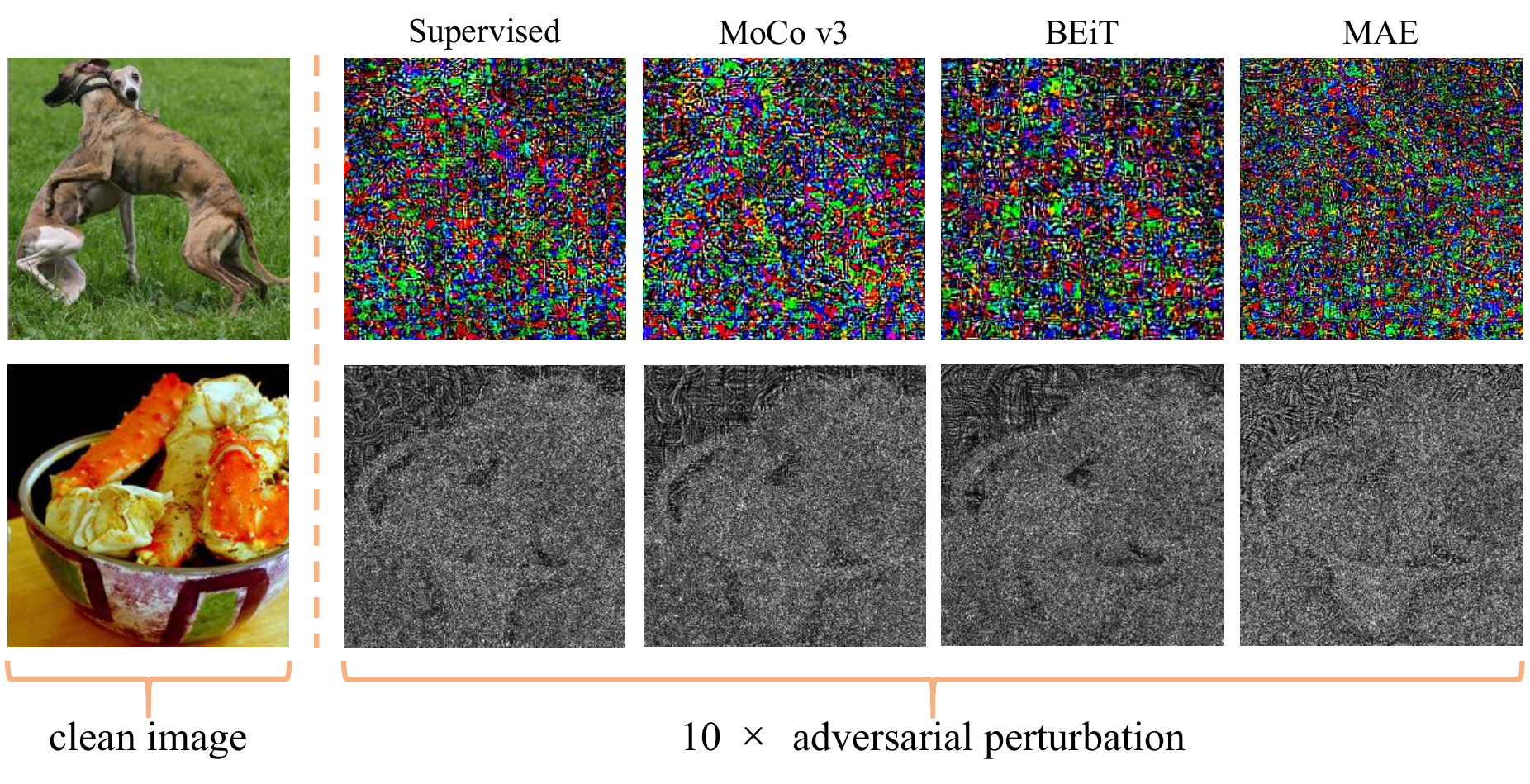}
\vspace{-1em}
\caption{Visualization of adversarial perturbations computed on ViT-B/16 under different training methods, including both RGB images and gray images. \textbf{MAE's perturbation is obviously more homogeneous across the spatial space than other methods.}}
\label{fig:residual}
\end{figure}

Some visualization results and similarity analysis tell us that, compared to other training methods, MAE is a more homogeneous vision learner. 
This unanticipated discovery hits us when we visualize the adversarial perturbations computed on vision transformers that are pretrained by different methods. 
Here we apply the $\ell_\infty$-norm PGD attack, where the generation of adversarial examples heavily rely on the back-propagation of gradient information. 
Based on the spatial distribution and the local intensity of adversarial perturbations, it is reasonable to understand which patches are given more attention by the model. 
We provide two kinds of visualization in \Fref{fig:residual}: 
1) the first row gives the raw RGB values of residuals; 
2) the second row gives the gray images of residuals that are averaged over raw RGB values. 
The brighter part indicates the greater salience.

\begin{figure*}[t]
\vspace{-1em}
\centering
\includegraphics[width=1.0\linewidth]{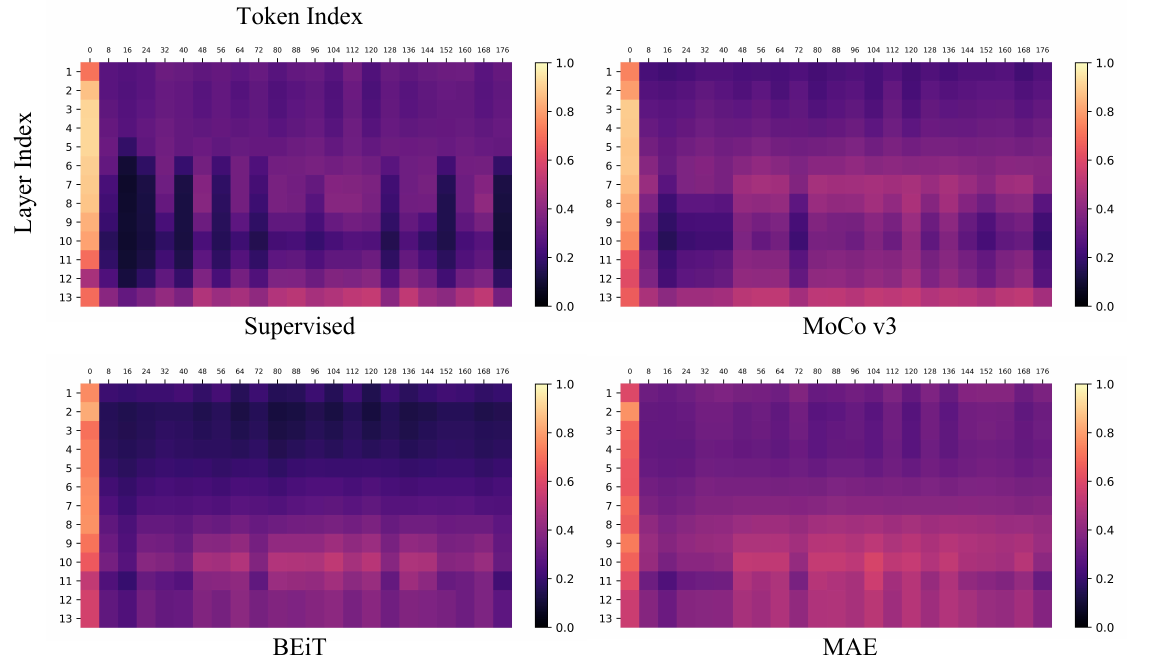}
\vspace{-2em}
\caption{Centered kernel alignment (CKA) similarity calculated between spatial tokens and class tokens, where token index ``0'' means the class token and layer index ``13'' means the normalized activation. The brighter lattice denotes the larger CKA similarity, while the darker lattice denotes the weaker CKA similarity. \textbf{The tokens of both BEiT and MAE show the great correlation in each transformer layer.}}
\label{fig:heatmap}
\end{figure*}

\begin{figure}[t]
\centering
\includegraphics[width=0.9\linewidth]{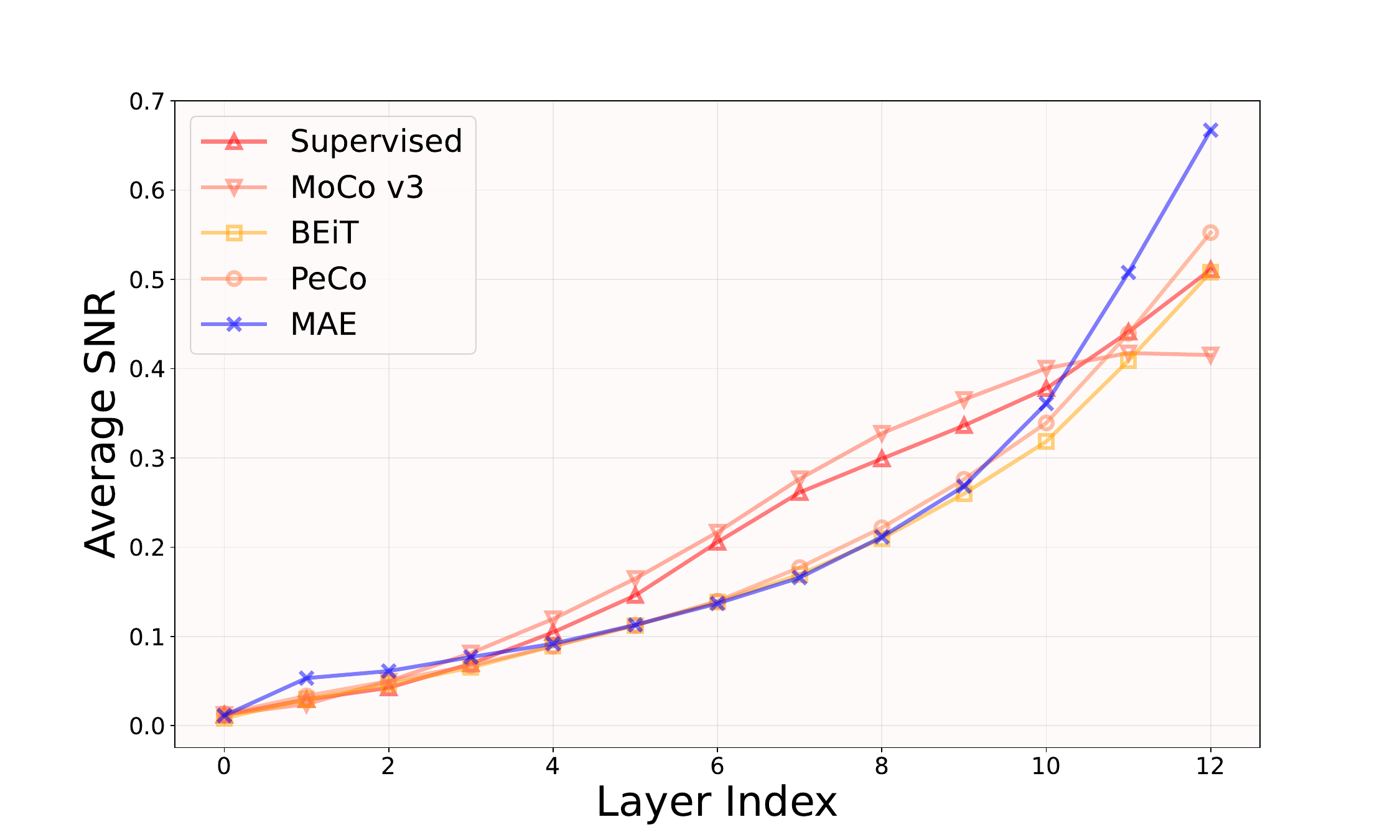}
\caption{Average signal to noise (SNR) ratios for exploring which layers does adversarial perturbation affect more on different methods, where the larger SNR indicates the more feature deviation. \textbf{MAE shows much larger representation deviation on the last three layers when encountering adversarial perturbation.}}
\label{fig:wa_snr}
\end{figure}

\begin{table}[t]
	\footnotesize
    \centering
    \setlength{\tabcolsep}{1mm}{
    \begin{tabular}{lp{12mm}<{\centering}p{15mm}<{\centering}p{9mm}<{\centering}p{9mm}<{\centering}p{9mm}<{\centering}}
        \toprule
        Method
        & Supervised
        & MoCo v3
        & BEiT
        & PeCo
        & MAE
        \\
        \midrule
        Var ($\times10^{-4}$) & 1.77 & 1.84 & 1.69 & 1.66 & \textbf{1.54}
        \\
        \bottomrule
    \end{tabular}
    }
    \vspace{0.5em}
    \caption{Average variance of adversarial perturbations on ViT-B/16 under different pretraining methods. \textbf{MAE's adversarial perturbation has the lowest variance among pixels.}}
    \label{tab:table2}
\end{table}

From the visual results, we can easily distinguish the salient part for image recognition according to adversarial perturbations computed on the supervised baseline and MoCo v3. 
By contrast, we can hardly figure out the salient part on MAE's perturbation. 
Furthermore, we calculate the average variance of these adversarial perturbations to quantitatively validate the observed vision homogeneity of MAE. 
From the results listed in Table \ref{tab:table2}, we find that MAE has the lowest perturbation variance among different pretraining methods. 
Additionally, we can see that vision BERT pretraining methods are overall more homogeneous than previous methods in visual understanding.

Another interesting point is illustrated in \Fref{fig:wa_snr}, where we intend to explore which layers are affected more by adversarial perturbations. 
To this end, we investigate the intermediate features in each layer of MAE encoder, including clean features that corresponds to clean input images and corrupted features from the corresponding adversarial samples. 
For each layer, we subsequently calculate the average signal to noise (SNR) value for each token between clean features and corrupted features, where the larger SNR value indicates the larger feature deviation. 
From \Fref{fig:wa_snr}, it is clear that all of three BERT pretraining methods have similar SNR except for the last three layers, where MAE shows significantly larger deviation. 
The reason of this phenomenon is probably MAE pretrained model changes more on the last three layers during the finetuning stage, due to the gap between pretraining knowledge from raw pixels and finetuning knowledge from semantics. We also observe that MAE often adopts larger layer-wise learning rate decay rate than BEiT and PeCo, thus larger learning rates are used for pretrained transformer blocks during finetuning.

To verify our claim that vision BERT pretraining results in the more homogeneous attention on different regions of images, we further conduct the experiments to calculate the similarity of token representations. 
Formally, given two representations or activations $\mathbf{X} \in \mathbb{R}^{k\times p_1}$ and $\mathbf{Y} \in \mathbb{R}^{k\times p_2}$ computed on $k$ images ($p_1$ and $p_2$ are the feature dimensions), we use Centered Kernel Alignment (CKA) \cite{CortesMR12,Kornblith0LH19} of these two representations to measure their normalized similarity. 
Let $\mathbf{K} = \mathbf{X}\mathbf{X}^\top$ and $\mathbf{L} = \mathbf{Y}\mathbf{Y}^\top$, we have
\begin{equation}
    \text{CKA} (\mathbf{K}, \mathbf{L}) = \frac{\text{HSIC}(\mathbf{K}, \mathbf{L})}{\sqrt{\text{HSIC}(\mathbf{K}, \mathbf{K})\text{HSIC}(\mathbf{L}, \mathbf{L})}}, 
\end{equation}
here $\text{HSIC}$ is Hilbert-Schmidt Independence Criterion \cite{GrettonFTSSS07}.

By calculating the CKA similarity between the class token and spatial tokens that are the output of each ViT self-attention block, we investigate the impacts contributed by different spatial tokens on the class token for the supervised baseline, MoCo v3, BEiT and MAE. 
We adopt the same ViT-B/16 backbone for these methods and test them on a randomly selected batch data (batch size is 256) from ImageNet-1k. 
With the class token as the reference, we can indirectly see the correlation between different spatial tokens. 
In  \Fref{fig:heatmap}, we uniformly sample the spatial token indexes and compute the CKA similarity between the class token and the spatial tokens denoted by these indexes. 
Apparently, for both the supervised baseline and MoCo v3, the correlation among their spatial tokens is weak, since their similarities between each other are really disparate. 
By contrast, the spatial tokens of BEiT and MAE show more similarities with the class token, especially in $7^{th}$, $8^{th}$ and $9^{th}$ ViT self-attention block. 
This result also reveals that vision BERT pretraining, especially MAE, is a more homogeneous learner on image context. Note that, considering PeCo shares a similar CKA map as BEiT, we have not shown its result in \Fref{fig:heatmap}.

\section{Boosting Adversarial Robustness of MAE}

In this section, we propose a visual prompting-based solution to alleviate MAE's weakness on adversarial robustness, \ie, learning a set of cluster-specific visual prompts in frequency-domain to assist the image recognition of MAE. When the prompts are successfully tuned, we directly append an additional prompting stage before we input images to the MAE model, requiring no model change.

\subsection{Frequency-Domain Visual Prompting}
Visual prompting \cite{jia2022vpt,bahng2022vp,huang2023damvp,bar2022visual} is an emerging topic that focuses to help large pretrained vision model \cite{radford2021learning,yuan2021florence,wang2022omnivl,weng2023transforming} adapt to downstream tasks through only changing the input space. Most of visual prompting methods consider their prompt design in the spatial domain of images. In our paper, it is the first time to explore the probability of prompting in the frequency domain of images. Formally, given a input image $\mathbf{x}$, we first convert it to frequency image $\mathbf{x}_f$ using Fast Fourier Transform (FFT). The prompted image $\mathbf{x}_p$ is derived by incorporating $\mathbf{x}_f$ with a learnable patch $\mathbf{p}$ and applying Inverse Fast Fourier Transform (IFFT), \ie, 
\begin{equation}
    \mathbf{x}_p \triangleq \mathop{\mathcal{T}_{ifft}}(\mathbf{x}_f + \mathbf{m}\cdot \mathbf{p}(\mathbf{x})),
\end{equation}
where $\mathcal{T}_{ifft}$ is the IFFT transformation and $\mathbf{p}(\mathbf{x})$ is the prompt selected for $\mathbf{x}$.  $\mathbf{m}$ denotes the mask that constrains the prompting in some particular frequency components (in our paper, we zero-out low-frequency areas with the radius of one-eighth the image size and preserve medium-/high-frequency areas by default, as shown in \Fref{fig:ours_pipeline}).

\subsection{Prompt Selection and Optimization}
Instead of using a generic prompt across the dataset, we choose to cluster the training data of MAE model\footnote{We use the checkpoint of pretrained ViT-B/16 provided in MAE official repository: \href{https://github.com/facebookresearch/mae}{https://github.com/facebookresearch/mae}.} into $N$ clusters ($N=1,000$ in our default setting) by Mini-Batch K-means algorithm \cite{sculley2010web} and learn a set of cluster-specific prompts following ``one prompt per cluster'' principle.
Formally, given the $N$ clusters and their corresponding prototypes $\{\mathbf{c}_i\}_{i=1}^N$, we can assign input image $\mathbf{x}$ with its frequency prompt $\mathbf{p}(\mathbf{x})$ by extracting the latent feature of $\mathbf{x}$ and searching the nearest prototype $\mathbf{c}_k$, \ie,\begin{equation}
    \mathbf{p}(\mathbf{x}) = \mathbf{p}_k, \quad\text{s.t.}\quad k = \mathop{\arg\min}_i \parallel \mathbf{E}(\mathbf{x}) - \mathbf{c}_i \parallel_2^2. 
\end{equation}

For prompt optimization, our goal is to encourage each prompt $\mathbf{p}_i$ to capture the common pattern of its corresponding cluster. 
Typically, we learn the prompt $\mathbf{x}_p$ by minimizing the cross-entropy loss, \ie, $\mathcal{L}_{CE} (\mathbf{x}_p, y)$, where $y$ is the ground-truth label of $\mathbf{x}_p$. 
By default, we adopt AdamW optimizer \cite{loshchilov2017decoupled} ($\beta_1=0.9$, $\beta_2=0.999$, $eps=10^{-8}$) with the base learning rate of $10^{-3}$ and the total batch size of 1,024. 
The whole procedure is implemented on 4 RTX A6000 GPUs for 100 epochs, where we use the first 5 epochs for warmup and decay the learning rate with half-cycle cosine after warmup. 
Intuitively, the strong commonality among the data within a single cluster reduces the difficulty of prompt optimization and facilitates the quick convergence.

During test-time verification, we found that the engagement of frequency prompts will partially harm the model performance on clean images, though the robustness is improved. To address this, we adopt the voting result as the final prediction, which ensembles the logits of both the input image $\mathbf{x}$ and the prompted image $\mathbf{x}_p$ as $\mathcal{M}(\mathbf{x}) + \lambda\mathcal{M}(\mathbf{x}_p)$. To balance the trade-off, we simply set $\lambda = 1$ by default.

\subsection{Robustness Evaluation}

We follow the same experimental setting in \Sref{sec:3.1}. Here we list several baseline methods for comparison: 

\textbf{Class-wise Adversarial Visual Prompts (C-AVP) \cite{chen2022visual}.} 
This solution devises class-wise visual prompts in the spatial domain of images, which are learned on adversarial examples to boost the robustness of classification models. Different from our method, it focuses on pixel-level prompts and learns one prompt per class in the dataset.

\begin{table}[t]
    \footnotesize
    \centering
    \setlength{\tabcolsep}{1mm}{
    \begin{tabular}{lcp{9mm}<{\centering}p{9mm}<{\centering}p{9mm}<{\centering}p{9mm}<{\centering}p{9mm}<{\centering}}
        \toprule
        \multicolumn{1}{c}{\multirow{2}{*}{Method}}
        & \multicolumn{1}{c}{\multirow{2}{*}{CA (\%)}} 
        & \multicolumn{5}{c}{Robust Acc (\%)} 
        \\
        \cmidrule{3-7}
        & 
        & PGD & MIM & BIM & C\&W & AA
        \\
        \midrule
        MAE & 83.6 & 1.9 & 3.6 & 1.9 & 5.9 & 0.0
        \\
        + C-AVP \cite{chen2022visual} & 50.2 & 31.3 & 34.0 & 30.7 & OOM & OOM
        \\
        + Perceptual & 83.9 & 9.3 & 15.1 & 9.1 & 14.6 & 2.3
        \\
        + ABP & 83.7 & 8.4 & 11.6 & 8.8 & 7.2 & 0.0
        \\
        \cellcolor{gray!20}\textbf{+ Ours$^\ast$} & \cellcolor{gray!20}\textbf{82.7} & \cellcolor{gray!20}\textbf{53.8} & \cellcolor{gray!20}\textbf{53.3} & \cellcolor{gray!20}\textbf{53.2} & \cellcolor{gray!20}\textbf{68.6} &
        \cellcolor{gray!20}\textbf{14.0}
        \\
        \cellcolor{gray!20}\textbf{+ Ours} & \cellcolor{gray!20}\textbf{82.7} & \cellcolor{gray!20}\textbf{69.3} & \cellcolor{gray!20}\textbf{69.2} & \cellcolor{gray!20}\textbf{69.3} & \cellcolor{gray!20}\textbf{56.8} & 
        \cellcolor{gray!20}\textbf{15.9}
        \\
        \midrule
        \textcolor{blue}{$\Delta$} & \textcolor{blue}{\textbf{-0.9}} & \textcolor{red}{\textbf{+67.4}} &
        \textcolor{red}{\textbf{+65.6}} &
        \textcolor{red}{\textbf{+67.4}} &
        \textcolor{red}{\textbf{+50.9}} & 
        \textcolor{red}{\textbf{+15.9}}
        \\
        \bottomrule
    \end{tabular}
    }
    \vspace{0.5em}
    \caption{Comparison of different robust methods on MAE. Here ``CA'' reports Top-1 accuracy on clean images. ``OOM'' is ``out of memory''. ``Our$^\ast$'' means using differentiable prompt selection. We unify $\epsilon=1/255$ for PGD, MIM, BIM, AA attack.}
    \label{tab:robustness_eval}
\end{table}

\begin{figure}[t]
\centering
\includegraphics[width=1\linewidth]{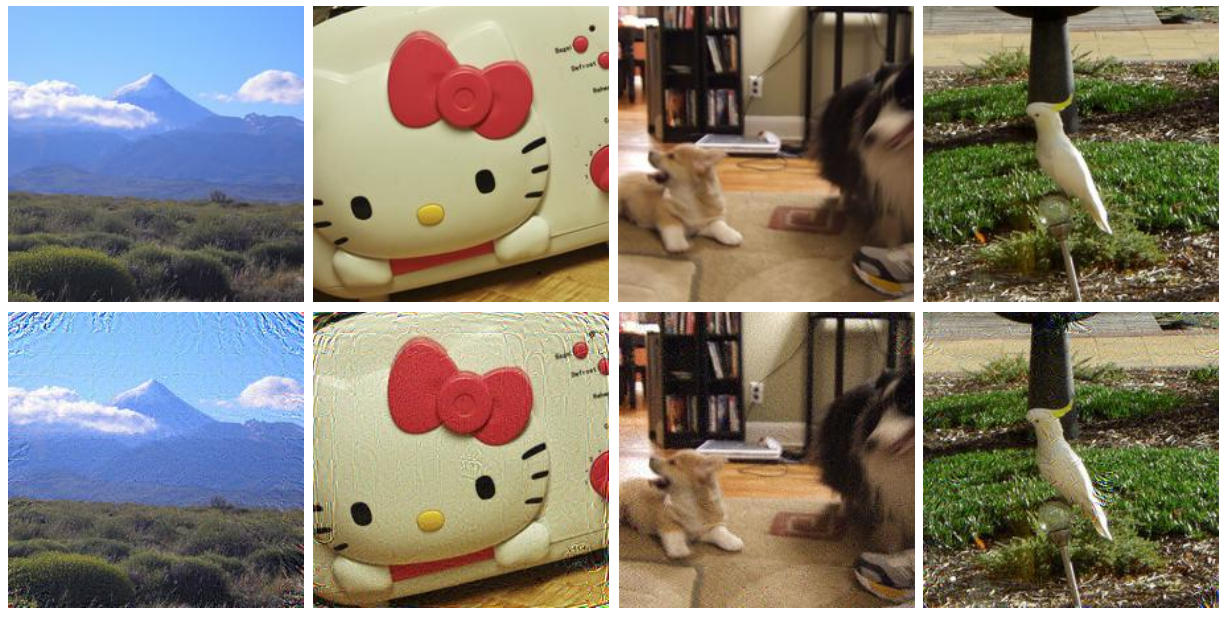}
\caption{Input images (\textbf{top}) and prompted images (\textbf{bottom}).}
\label{fig:image_vis}
\end{figure}

\textbf{Integrating Perceptual Loss}. 
This solution is to add an extra perceptual loss as the reconstruction supervision based on the analysis given in \Sref{sec:3.2}, thus encouraging MAE to reduce the dependency on high-frequency signals and comprehend more contextual information when learning the target semantics. This shares similar spirit to PeCo \cite{peco21}, but different from it, we directly use perceptual loss as BERT pretraining supervision rather than for VQVAE tokenizer training. To realize this, we leverage a pretrained ViT-tiny to compute the multi-layer feature distance between MAE decoded tokens and original image tokens in the masked region, so-called perceptual loss \cite{JohnsonAF16}. 
Here we empirically choose the features of $3^{rd}, 6^{th}, 9^{th}, 12^{th}$ transformer block, and the final perceptual loss is the sum of feature loss of these four blocks. The final pretraining objective loss is the weighted sum of the original pixel-level mean square error loss $L_{pix}$ and perceptual loss $L_{perc}$:
\begin{equation}
    L_{total} = L_{pix} + \mu * L_{perc}. 
\end{equation}
where $\mu=1$ by defualt. More details are in supplementary.

\textbf{Adversarial BERT pretraining (ABP).} This solution is to naturally integrate adversarial training \cite{madry2017towards} into BERT pretraining, which we call ``adversarial BERT pretrainin''. Specifically, we generate online adversarial perturbation upon unmasked patches by maximizing the reconstruction loss for masked patches, and conduct BERT pretraining with such online adversarial examples in a adversarial min-max manner. Similar to conventional adversarial learning, our intuition is to force the transformer model to learn on corrupted images that are adversarially generated by gradient-based attack in an online way. The difference is that we want to increase the robustness of pretraining rather than the supervised trained model. Additionally, we do not have any label information during BERT pretraining to guide the adversarial example generation. More details can be found in our supplementary.

\begin{table}[t]
    \footnotesize
    \centering
    \setlength{\tabcolsep}{0.6mm}{
    \begin{tabular}{llcp{8mm}<{\centering}p{8mm}<{\centering}p{8mm}<{\centering}p{8mm}<{\centering}p{8mm}<{\centering}}
        \toprule
        \multicolumn{1}{l}{\multirow{2}{*}{Dataset}} & \multicolumn{1}{l}{\multirow{2}{*}{Method}}
        & \multicolumn{1}{c}{\multirow{2}{*}{CA (\%)}} 
        & \multicolumn{5}{c}{Robust Acc (\%)} 
        \\
        \cmidrule{4-8}
        & & & PGD & MIM & BIM & C\&W & AA
        \\
        \midrule
        \multicolumn{1}{l}{\multirow{5}{*}{DTD}} & MAE & \textbf{72.4} & 39.4 & 39.8 & 39.8 & 43.4 & 5.2
        \\
        & + C-AVP & 46.0 & 55.1 & 53.8 & 55.0 & OOM & 25.4
        \\
        & + Perceptual & 75.9 & 42.6 & 43.1 & 42.7 & 49.4 & 11.2
        \\
        & + ABP & 70.4 & 41.9 & 45.6 & 41.6 & 37.5 & 9.1
        \\
        & \cellcolor{gray!20}\textbf{+ Ours} & \cellcolor{gray!20}72.2 & \cellcolor{gray!20}\textbf{58.2} & \cellcolor{gray!20}\textbf{59.8} & \cellcolor{gray!20}\textbf{59.6} & \cellcolor{gray!20}\textbf{65.9} & \cellcolor{gray!20}\textbf{32.0} 
        \\
        \midrule
        \multicolumn{1}{l}{\multirow{5}{*}{CUB200}} & MAE & 81.8 & 17.8 & 18.1 & 18.2 & 23.4 & 0.2
        \\
        & + C-AVP & 43.5 & 39.1 & 38.8 & 39.0 & OOM & 12.7
        \\
        & + Perceptual & 82.1 & 19.0 & 22.3 & 19.3 & 25.5 & 3.9
        \\
        & + ABP & 78.0 & 24.3 & 26.1 & 24.5 & 35.8 & 5.4
        \\
        & \cellcolor{gray!20}\textbf{+ Ours} & \cellcolor{gray!20}\textbf{81.9} & \cellcolor{gray!20}\textbf{62.8} & \cellcolor{gray!20}\textbf{66.8} & \cellcolor{gray!20}\textbf{64.8} & \cellcolor{gray!20}\textbf{64.6} & \cellcolor{gray!20}\textbf{23.0} 
        \\
        \bottomrule
    \end{tabular}
    }
    \vspace{0.5em}
    \caption{Comparison of different robust methods on MAE. Here ``CA'' reports Top-1 accuracy on clean images. ``OOM'' is ``out of memory''. We unify $\epsilon=4/255$ for PGD, MIM, BIM, AA attack.}
    \label{tab:other1}
\end{table}

\begin{table}[t]
    \footnotesize
    \centering
    \setlength{\tabcolsep}{0.6mm}{
    \begin{tabular}{lcp{7mm}<{\centering}p{7mm}<{\centering}p{7mm}<{\centering}p{7mm}<{\centering}p{7mm}<{\centering}}
        \toprule
        \multicolumn{1}{l}{\multirow{2}{*}{Method}}
        & \multicolumn{1}{c}{\multirow{2}{*}{Clean Acc (\%)}} 
        & \multicolumn{5}{c}{Robust Acc (\%)} 
        \\
        \cmidrule{3-7}
        & 
        & PGD & MIM & BIM & C\&W & AA
        \\
        \midrule
        Supervised & 81.8 & 8.8 & 12.2 & 8.9 & 18.6 & 0.1
        \\
        \cellcolor{gray!20}\textbf{+ Ours} &
        \cellcolor{gray!20}\textbf{81.1} & \cellcolor{gray!20}\textbf{45.0} & \cellcolor{gray!20}\textbf{45.6} &  \cellcolor{gray!20}\textbf{44.9} & \cellcolor{gray!20}\textbf{60.1} & \cellcolor{gray!20}\textbf{11.9}
        \\
        \midrule
        BEiT & 83.2 & 6.9 & 9.8 & 6.9 & 21.6 & 0.0
        \\
        \cellcolor{gray!20}\textbf{+ Ours} & \cellcolor{gray!20}\textbf{83.4} & \cellcolor{gray!20}\textbf{30.7} & \cellcolor{gray!20}\textbf{36.5} & \cellcolor{gray!20}\textbf{30.4} & \cellcolor{gray!20}\textbf{58.7} & \cellcolor{gray!20}\textbf{5.7}
        \\
        \midrule
        PeCo & 84.5 & 13.1 & 15.9 & 13.3 & 24.0 & 0.1
        \\
        \cellcolor{gray!20}\textbf{+ Ours} & \cellcolor{gray!20}\textbf{83.9} & \cellcolor{gray!20}\textbf{35.4} & \cellcolor{gray!20}\textbf{37.5} & \cellcolor{gray!20}\textbf{35.3} & \cellcolor{gray!20}\textbf{56.3} & \cellcolor{gray!20}\textbf{8.9}
        \\
        \bottomrule
    \end{tabular}
    }
    \vspace{1em}
    \caption{Results of our method on other pretraining paradigms.}
    \label{tab:other2}
\end{table}

\Tref{tab:robustness_eval} showcases the robustness of MAE models when equipped with different training strategies. 
For fair comparison with baseline method C-AVP, our default setting uses $\arg\min$ to select the prompt, so the gradient is only calculated within the classification branch. 
However, the gradient can not be back-propagated through $\arg\min$ operation, thus we replace the non-differentiable $\arg\min$ with a differentiable gumble-softmax \cite{jang2016categorical} sampling and re-evaluate the performance. 
From \Tref{tab:robustness_eval}, our method achieves the clear improvement on adversarial robustness while keeping limited clean accuracy sacrificed. 
Since our gumble-softmax version still achieves strong robustness (even better than the default version on C\&W, it can also prove that the robustness of our method is not from the gradient obfuscated. 
Our visual prompts fulfill the medium-/high-frequency of images with the learned domain patterns, thus reducing the redundancy of these frequency components and narrowing the optimization space of adversarial perturbations.

\subsection{Additional Results}

\textbf{More Datasets. }
We further validate the proposed method on two classification datasets, \ie, DTD \cite{dtd} and CUB200 \cite{cub200}. 
DTD is a dataset that collects 47 kinds of different textures and CUB200 contains images of various birds. 
On such datasets, visual prompts are expected to capture the more fine-grained representations than that are on ImageNet-1k. 
As shown in \Tref{tab:other1}, our method keep the superiority on these fine-grained classification tasks, with its great improvement on robust accuracy.

\textbf{More Pretraining Methods.} 
We further apply our method on supervised pretraining, BEiT and PeCo. 
\Tref{tab:other2} shows that our method can also works for other pretraining paradigms but gains relatively less on MAE, \ie, our medium-/high-frequency visual prompt better fits MAE.

\begin{table}[t]
    \footnotesize
    \centering
    \setlength{\tabcolsep}{1mm}{
    \begin{tabular}{lcp{8mm}<{\centering}p{8mm}<{\centering}p{8mm}<{\centering}p{8mm}<{\centering}p{8mm}<{\centering}}
        \toprule
        \multicolumn{1}{c}{\multirow{2}{*}{Setting}}
        & \multicolumn{1}{c}{\multirow{2}{*}{CA (\%)}} 
        & \multicolumn{5}{c}{Robust Acc (\%)} 
        \\
        \cmidrule{3-7}
        & 
        & PGD & MIM & BIM & C\&W & AA
        \\
        \midrule
        \multicolumn{6}{l}{\textit{spatial domain vs. frequency domain}}
        \\
        full spatial. & 78.3 & 10.5 & 11.6 & 10.4 & 16.3 & 0.9
        \\
        \cellcolor{gray!20}full freq. & \cellcolor{gray!20}80.5 & \cellcolor{gray!20}26.6 & \cellcolor{gray!20}28.1 & \cellcolor{gray!20}26.7 & \cellcolor{gray!20}19.7 & \cellcolor{gray!20}3.2
        \\
        \midrule
        \multicolumn{6}{l}{\textit{universal prompt vs. cluster-specific prompt}}
        \\
        universal & 78.3 & 35.2 & 35.9 & 35.3 & 24.0 & 3.8
        \\
        \cellcolor{gray!20}cluster-specific & \cellcolor{gray!20}82.7 & \cellcolor{gray!20}69.3 & \cellcolor{gray!20}69.2 & \cellcolor{gray!20}69.3 & \cellcolor{gray!20}56.8 & \cellcolor{gray!20}15.9
        \\
        \midrule
        \multicolumn{6}{l}{\textit{masking low frequency vs. masking high frequency}}
        \\
        mask high-freq. & 83.7 & 2.6 & 6.0 & 2.7 & 4.8 & 0.0
        \\
        \cellcolor{gray!20}mask low-freq. & \cellcolor{gray!20}82.7 & \cellcolor{gray!20}69.3 & \cellcolor{gray!20}69.2 & \cellcolor{gray!20}69.3 & \cellcolor{gray!20}56.8 & \cellcolor{gray!20}15.9
        \\
        \midrule
        \multicolumn{6}{l}{\textit{whether using logit ensemble}}
        \\
        w/o ensemble & 80.5 & 70.3 & 70.4 & 70.4 & 51.8 & 16.1
        \\
        \cellcolor{gray!20}w/ ensemble & \cellcolor{gray!20}82.7 & \cellcolor{gray!20}69.3 & \cellcolor{gray!20}69.2 & \cellcolor{gray!20}69.3 & \cellcolor{gray!20}56.8 & \cellcolor{gray!20}15.9
        \\
        \midrule
        \multicolumn{6}{l}{\textit{Number $N$ of clusters}}
        \\
        $N=500$ & 81.3 & 68.6 & 68.8 & 68.8 & 53.1 & 12.2
        \\
        $N=1,500$ & 82.3 & 62.4 & 63.0 & 62.7 & 47.1 & 9.2
        \\
        \cellcolor{gray!20}$N=1,000$ & \cellcolor{gray!20}82.7 & \cellcolor{gray!20}69.3 & \cellcolor{gray!20}69.2 & \cellcolor{gray!20}69.3 & \cellcolor{gray!20}56.8 & \cellcolor{gray!20}15.9
        \\
        \bottomrule
    \end{tabular}
    }
    \vspace{1em}
    \caption{Ablation studies to verify the effect of different designs. The highlight results corresponds to our default setting.}
    \label{tab:ablation_1}
\end{table}

\subsection{Ablation Study}

\Fref{fig:image_vis} showcases some instances of our prompted images. It indicates that though our experiments is focused on classification, our frequency-domain prompting has great potentials to well perform in other tasks like image segmentation or detection, since it does not use a visual-explicit way like spatial-domain patches used in \cite{bahng2022vp} and rarely influences the visual quality. 
In \Tref{tab:ablation_1}, we also give the ablation studies to prove the validity about some of our designs: 
1) we observe that prompting on the frequency domain obtains higher robust accuracy than that is on the spatial domain, frequency-domain prompting is more effective for boosting MAE's robustness; 
2) we show that learning cluster-specific prompts (``one cluster one prompt'') works better than learning a single prompt for all of images (``universal prompt''); 
3) we show the validity of medium-/high-frequency prompting (\ie, masking on low frequency in \Fref{fig:ours_pipeline}) fits MAE better than low-/medium-frequency prompting (``masking high-freq.''); 
4) we try to discard our logit ensemble and find the minor gains on robust accuracy but non-negligible degradation on clean accuracy, proving that the ensemble is necessary to benefit the clean performance; 
5) we ablate the number $N$ of clusters and choose $N=1,000$ in our experiment considering both robustness and clean performance.

\section{Conclusions}
This paper analyzes the adversarial robustness of vision BERT pretraining. We observe that the main design difference of existing vision BERT pretraining methods, \ie, BERT pretraining target designs, is the key impact factor. More specifically, adopting the raw pixels of masked patches as the BERT pretraining target (\eg, MAE) degrades the adversarial robustness of the model than the counterparts predicting the semantic context (\eg, BEiT, PeCo). Through the detailed analysis, we hypothesize that directly predicting raw pixels makes the MAE model more sensitive to medium-/high-frequency signals of images. To improve the robustness of MAE, a simple yet effective solution have been proposed based on our analysis, \ie, learning frequency-domain visual prompts that contains dataset-extracted knowledge to assist the test-time model prediction. We surprisingly find that this solution can greatly boost the adversarial robustness of MAE while maintaining its performance on clean samples. We hope that the analysis of our paper can motivate more study in this direction.

\noindent\textbf{Acknowledgement.} 
This work was supported in part by the Natural Science Foundation of China under Grant  U20B2047, 62072421, 62002334, 62102386 and 62121002, and by Fundamental Research Funds for the Central Universities under Grant WK2100000011.

\section{Supplementary Materials}

\subsection{Integrating Perceptual Loss in MAE Pretraining}

With the analysis given in \Sref{sec:3.2}, it is clear that MAE relies more on medium-/high-frequency signals of input images due to only predicting raw pixels. 
In the light of BEiT \cite{beit21} and PeCo \cite{peco21}, a natural idea for improving MAE's ability on adversarial robustness is involving some semantic context in reconstruction rather than only moving towards raw image pixels. 
To realize this, we leverage a pretrained ViT-Tiny to compute the multi-layer feature distance between MAE decoded image and original image tokens in the masked region, which is so-called perceptual loss \cite{JohnsonAF16}. Formally, Let $f_l (x)$ be the normalized feature of $l$-th transformer block of the pretrained ViT-tiny model. Then for the original image $x$ and the reconstructed image $\hat{x}$, we can formulate the perceptual loss as:
\begin{equation}
    L_{perc} = \sum_{l\in \{3,6,9,12\}}\parallel f_l (x) - f_l (\hat{x}) \parallel_2^2, 
\end{equation}
Here we empirically use the feature of $3^{rd}, 6^{th}, 9^{th}, 12^{th}$ transformer block, and the final perceptual loss is the sum of feature loss of these four blocks. 
The final pretraining objective loss is the sum of the original pixel-level mean square error loss $L_{pix}$ and perceptual loss $L_{perc}$:
\begin{equation}
    L_{total} = L_{pix} + \lambda * L_{perc}. 
\end{equation}
where $\lambda$ is set as 1 by default. With this simple design, MAE can reduce its dependence on high-frequency signal of images while inheriting original pixel reconstruction loss to maintain the great learning ability. Note that, this improvement only brings limited extra training budget since we just apply a ViT-Tiny model for perceptual loss computation, which is efficient enough during forward or backward compared to the total training time. 

To validate the effectiveness of improved MAE that is equipped with our multi-layer perceptual loss, we test its adversarial robustness under the same setting as given in \Tref{tab:robustness_eval}. According to the quantitative results showcased, the involvement of perceptual loss can comprehensively boost the ability of MAE. 
Both clean classification accuracy and adversarial robustness are improved to a nearly satisfying level.

\begin{table*}[t]
	\footnotesize
    \centering
    \setlength{\tabcolsep}{0.35mm}{
    \begin{tabular}{lc|p{12mm}<{\centering}p{12mm}<{\centering}p{12mm}<{\centering}p{12mm}<{\centering}|p{12mm}<{\centering}p{12mm}<{\centering}p{12mm}<{\centering}p{12mm}<{\centering}}
        \toprule
        \multicolumn{1}{c}{\multirow{2}{*}{Method}}
        & \multicolumn{1}{c}{\multirow{2}{*}{Clean Acc (\%)}} 
        & \multicolumn{4}{|c}{Robust Acc (\%) ($\varepsilon=0.001$, $\eta=0.2$)} 
        & \multicolumn{4}{|c}{Robust Acc (\%) ($\epsilon=0.005$, $\eta=0.3$)} 
        \\
        \cmidrule{3-6}
        \cmidrule{7-10}
        & 
        & PGD\cite{Madry18adversarial}
        & MIM\cite{dong18mifgm}
        & BIM\cite{kurakin2016adversarial}
        & AA\cite{croce2020autoattack}
        & PGD\cite{Madry18adversarial}
        & MIM\cite{dong18mifgm}
        & BIM\cite{kurakin2016adversarial}
        & AA\cite{croce2020autoattack}
        \\
        \midrule
        Supervised & 81.8 & 58.7 & 39.3 & 55.8 & 13.7 & 0.2 & 1.1 & 20.8 & 4.3
        \\
        MoCo v3 \cite{Chen21mocov3} & 83.1 & 61.1 & 46.4 & 59.9 & 20.9 & 0.2 & 1.7 & 21.5 & 7.8
        \\
        BEiT \cite{beit21} & 83.2 & 64.8 & 40.8 & 53.9 & 19.2 & 4.6 & 0.7 & 15.8 & 6.8
        \\
        PeCo \cite{peco21} & 84.5 & 64.7 & 41.5 & 56.4 & 15.9 & 3.8 & 1.2 & 21.8 & 4.9
        \\
        MAE \cite{mae21} & 83.6 & \textbf{58.4} & \textbf{28.9} & \textbf{46.3} & \textbf{5.9} & \textbf{0.1} & \textbf{0.3} & \textbf{11.6} & \textbf{1.3}
        \\
        \bottomrule
    \end{tabular}
    }
    \vspace{0.5em}
    \caption{Comparison of robustness to various $\ell_2$-norm adversarial attacks among different BERT training methods.  All methods are equally implemented on ViT-Base backbone, and MAE shows significantly worse robustness than other methods. Here, ``Clean Acc" and ``Robust Acc" mean the classification accuracy on clean samples and adversarial samples respectively. For PGD, BIM and MIM, we set the budget as $\epsilon=\varepsilon*\sqrt{C\times H\times W}$, where $C,H,W$ denotes the dimensions of the input image. For AA, we set the budget as $\eta$. }
    \label{tab:table1}
\end{table*}

\subsection{Adversarial BERT pretraining}

Besides the first baseline, we also consider another intuitive baseline, \ie, introduce adversarial learning into MAE pretraining to improve its robustness. Similar to conventional adversarial learning, our intuition is to force the transformer model to learn on corrupted images that are adversarially generated by gradient-based attack in an online way. The difference is that we want to increase the robustness of pretraining rather than the supervised trained model. Subsequently, we do not have any label information during BERT pretraining to guide the adversarial sample generation.

\begin{algorithm}[t]
\caption{Adversarial BERT Pretraining}
\SetAlgoLined
    \KwIn{clean image set $\boldsymbol{X}$; transformer encoder $\mathbf{E}$; auxiliary prediction head $\mathbf{h}$; pixel reconstruction loss $L_{pix}$}
    \KwOut{Parameters $\mathbf{\theta}$ of $\mathbf{E}$ and $\mathbf{h}$;}
    \For{each sampled mini-batch $\{\mathbf{x}\} \in \boldsymbol{X}$}{ 
        $\circ$ Generate random masks $\{\mathbf{m}\}$ for $\{\mathbf{x}\}$ \\
        $\circ$ Forward masked mini-batch $\{\mathbf{x}_{\mathbf{m}}\}$ to get reconstructed results $\{\hat{\mathbf{x}}\}$: \\
           \qquad $\hat{\mathbf{x}} = \mathbf{h}(\mathbf{E}(\mathbf{x}_{\mathbf{m}})) $ \\
        $\circ$ Generate the adversarial mini-batch $\{\mathbf{x}_{adv}\}$ with stand untargeted $\ell_\infty$ PGD attack: \\
        \For{t=1,$\cdots$,T}{
         $\mathbf{x}^t_{adv} = \mathbf{x}_{adv}^{t-1} + \epsilon \nabla_{\mathbf{x}_{adv}^{t-1}} [\mathbf{m} \cdot L_{pix} (\mathbf{x}, \hat{\mathbf{x}}_{adv}^{t-1})]$ \\
        }
        $\circ$ Feed the adversarial mini-batch $\{\mathbf{x}_{adv}\}$ into pretraining by minimizing: \\
        \qquad $L_{pix}(\mathbf{x}, \hat{\mathbf{x}}_{adv})$
    } 
\label{alg:adv_bert}
\end{algorithm}

As shown in Algorithm \ref{alg:adv_bert}, when integrating adversarial BERT pretraining with MAE, we follow the original MAE pretraining procedure and add the online generated adversarial samples into training alternatively.  Since MAE encoder only operates on unmasked tokens, we only add adversarial perturbations onto the unmasked regions. Following untargeted standard $\ell_\infty$-norm PGD attack, we iteratively generate adversarial masked images by maximizing the MAE pixel reconstruction loss, \ie, 
\begin{equation}
    \mathbf{x}^t_{adv} = \mathbf{x}_{adv}^{t-1} + \epsilon \nabla_{\mathbf{x}_{adv}^{t-1}} [\mathbf{m} \cdot L_{pix} (\mathbf{x}, \hat{\mathbf{x}}_{adv}^{t-1})], \quad t \in [1,T], 
\end{equation}
where $\epsilon$ denotes the attack step size, the superscript $\hat{}$ denotes the reconstructed image, $\mathbf{m}$ is the generated random mask for BERT pretraining, and $T$ is the total iteration number and set as 4 by default. 
After generating the online adversarial samples $\mathbf{x}_{adv}$, we will add them into MAE pretraining as hard samples and predict the groundtruth raw pixel values from clean samples in an adversarial way. 
\begin{equation}
    \mathop{\text{min}}_\theta L_{pix} (\mathbf{x}, \hat{\mathbf{x}}_{adv}).
\end{equation}

\begin{figure*}[h]
\centering
\includegraphics[width=1.0\linewidth]{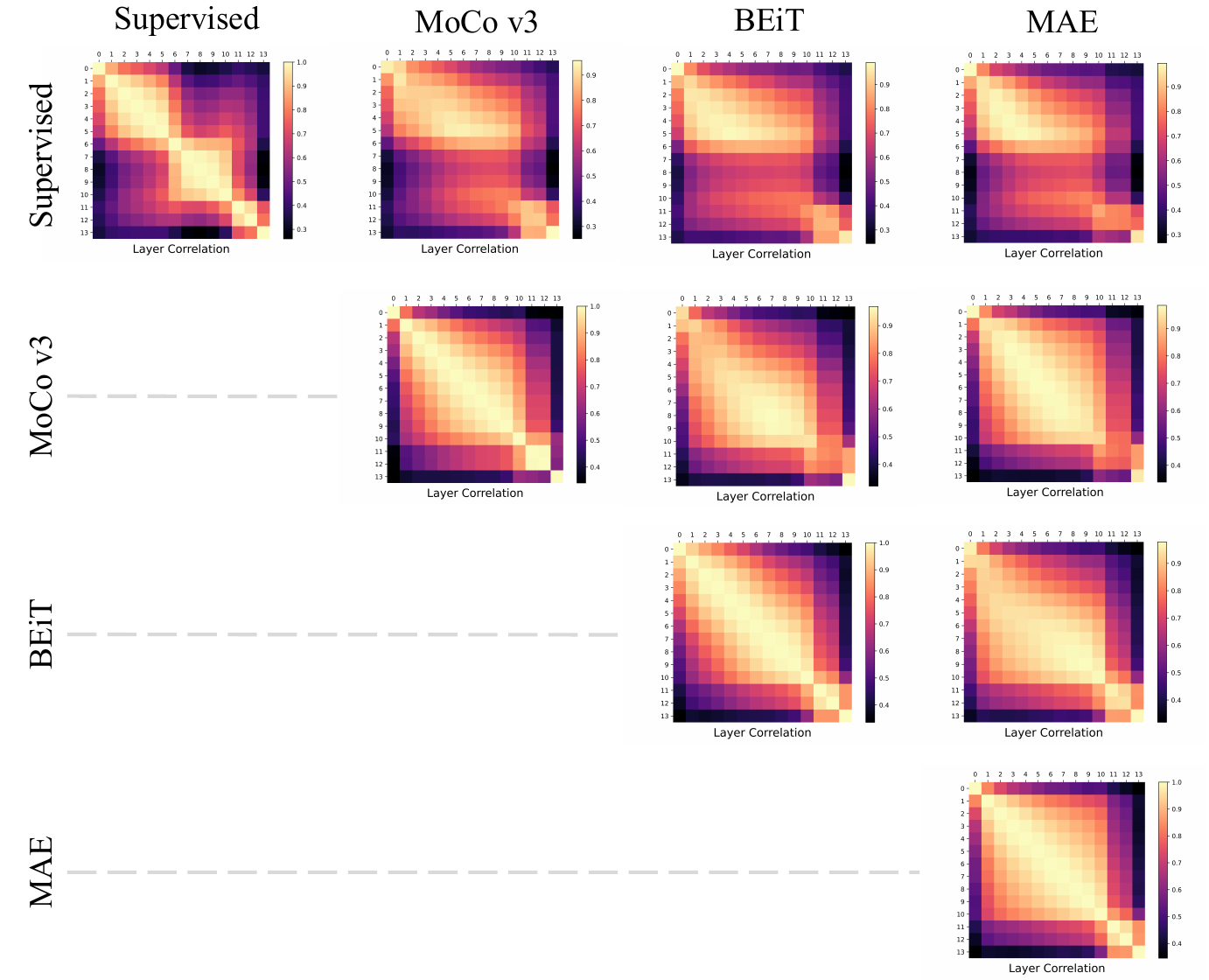}
\caption{Centered kernel alignment (CKA) similarity calculated on the layer outputs of ViTs that are equipped with different training methods, where ``1-12'' denotes the different ViT block layer and the layer index ``13'' means the normalized activation. The brighter lattice denotes the larger CKA similarity, while the darker lattice denotes the weaker CKA similarity.}
\label{fig:layer_correlation}
\end{figure*}

Likewise, we conduct the robustness evaluation of MAE over the same setting after equipping with adversarial BERT pretraining. From the results shown in \Tref{tab:robustness_eval}, we surprisingly find that both clean classification accuracy and adversarial robustness are boosted. 
It is a counterintuitive phenomenon, since adversarial training is usually a trade-off game to balance the clean performance and adversarial robustness with proper harness, \ie, one of them can be boosted while the other one degrades. Especially, we find the robustness improvement on C\&W and Auto-Attack are not that significant, which may be because C\&W and Auto-Attack share different adversarial perturbation generation mechanisms from PGD. This adversarial BERT pretraining gives a successful try to comprehensively improve MAE's ability and will inspire more research works in exploring better transferable adversarial BERT pretraining in this direction.

\subsection{Robustness Evaluation on $\ell_2$-Norm Attack}

We provide the adversarial robustness results to $\ell_2$-norm attack for ViTs that are equipped with different training methods, including the supervised baseline, MoCo v3 \cite{Chen21mocov3}, BEiT \cite{beit21}, PeCo \cite{peco21} and MAE \cite{mae21}. 
Specifically, we implement the $\ell_2$ version of PGD \cite{Madry18adversarial}, BIM\cite{kurakin2016adversarial}, MIM \cite{dong18mifgm} and AA\cite{croce2020autoattack} attacks for the verification. 
For PGD, BIM and MIM, we set the budget as $\epsilon=\varepsilon*\sqrt{C\times H\times W}$, where $C,H,W$ denotes the dimensions of the input image. 
For AA, we set the budget as $\eta$. 
As the results given in Table \ref{tab:table1}, MAE's robustness degradation is also significant, further demonstrating the interesting observation raised in our manuscript. 
\textbf{Overall, MAE has obviously degradation on adversarial robustness when compared with other vision BERT pretraining methods.}

\subsection{Layer Correlation}

We further provide the layer correlation results for different vision training methods. 
To better understand the similarity of learned representations (\ie, layer outputs) of these training methods, we tested both 
1) the CKA results between different layers of themselves and 
2) the CKA results across different layers from different trained ViTs.

From \Fref{fig:layer_correlation}, we can easily find that \textbf{vision BERT pretraining methods enable the ViT model having the stronger correlation between lower layers and higher layers like previous pretraining methods}, while the ViT trained in supervision shows much worse layer correlation by contrast. 

Besides, it can be observed that \textbf{different methods share the similar shallow layer representations}. 
The reason might be that, as we speculated, they follow the similar way to extract the local information from input images before the feature fusion in the higher layers.


{\small
\bibliographystyle{ieee_fullname}
\bibliography{egbib}
}

\end{document}